\documentclass[opre]{informs}
\OneAndAHalfSpacedXI
\usepackage{natbib}
\bibpunct[, ]{(}{)}{,}{a}{}{,}%
\def\bibfont{\small}%
\usepackage{tikz}
\usepackage{booktabs} 
\usepackage{framed}
\usepackage{mathtools}
\usepackage{comment}
\usepackage[normalem]{ulem}
\usepackage{enumitem}
\usepackage{amsmath}
\usepackage{epstopdf}
\usepackage{graphicx}
\usepackage{threeparttable}
\usepackage{subcaption}
\usepackage{svg} % 需使用包
\usepackage{multirow}
\usepackage{threeparttable}
\usepackage{url}
\usepackage{subfloat}
\usepackage{bbm}
\usepackage{longtable}
\usepackage{hyperref}
\usepackage{diagbox}
\usepackage{tablefootnote}
\usepackage[justification=centering]{caption}
\usepackage{graphicx}
\usepackage{epstopdf}
\graphicspath{{figures/}}
\newcommand{\outline}[1]{%
    \tikz[baseline=(text.base)]{
        \node[anchor=base,draw=black,thick,outer sep=0.5pt,fill=white] (text) {#1};
    }%
}

\hypersetup{
	colorlinks=true,
	linkcolor=black,
	citecolor=blue
} 

\usepackage[hyphenbreaks]{breakurl}
\definecolor{myred}{rgb}{0.75,0.0,0.0}

\usepackage{stackengine}

\usepackage{algorithm}
\usepackage[noend]{algpseudocode}

\usepackage{fancybox}
\usepackage[framemethod=tikz]{mdframed}
\usepackage{lipsum}
\usetikzlibrary{shadows}
\newmdenv[shadow=true,shadowcolor=black]{myshadowbox}

\TheoremsNumberedThrough % Preferred (Theorem 1, Lemma 1, Theorem 2)
\ECRepeatTheorems

\EquationsNumberedThrough % Default: (1), (2),...
\MANUSCRIPTNO{MS-XXXX-XXXX.XX}
\makeatletter
\usepackage[utf8]{inputenc} % allow utf-8 input
\usepackage[T1]{fontenc}    % use 8-bit T1 fonts
\usepackage{hyperref}       % hyperlinks
\usepackage{url}            % simple URL typesetting
\usepackage{booktabs}       % professional-quality tables
\usepackage{amsfonts}       % blackboard math symbols
\usepackage{nicefrac}       % compact symbols for 1/2, etc.
\usepackage{microtype}      % microtypography
\usepackage{amsmath}         % colors

\usepackage{graphicx}
\usepackage{multirow}
\usepackage{pifont}
\usepackage{makecell}
\usepackage{colortbl}

\usepackage{xcolor}
\definecolor{mygray}{RGB}{192,192,192}
\definecolor{mygray2}{RGB}{230,230,230}
\usepackage[most]{tcolorbox}
\usepackage{float}
\usepackage{xspace}
\tcbset{
  aibox/.style={
    width=\textwidth,
    top=10pt,
    colback=white,
    colframe=black,
    colbacktitle=black,
    enhanced,
    center,
    attach boxed title to top left={yshift=-0.1in,xshift=0.15in},
    boxed title style={boxrule=1pt,colframe=black,colback=mygray2, colbacktitle=black},
    coltitle = black,
    fonttitle=\bfseries, % 加粗标题字体
  }
}
\newtcolorbox{AIbox}[2][]{aibox,title=#2,#1}
\usepackage{listings}
\definecolor{myblue}{RGB}{100, 150, 200}
\definecolor{mygreen}{RGB}{80, 160, 80}

\definecolor{darkgreen}{rgb}{0.0, 0.5, 0.0}
\definecolor{darkgray}{gray}{0.4}
\definecolor{maroon}{rgb}{0.5, 0.0, 0.0}
\definecolor{navy}{rgb}{0.0, 0.0, 0.5}
\definecolor{teal}{rgb}{0.0, 0.5, 0.5}

\usepackage{wrapfig,lipsum,booktabs}
\newtheorem{proper}{Requirement}%[section]
\usepackage{tikz}
\usetikzlibrary{shapes, arrows.meta, positioning}

\begin{document}
	%%%%%%%%%%%%%%%%
	
	% default values will be used.
	%\setcounter{page}{1}
	%\DOI{10.1287/xxxx.0000.0000}%
	
	\RUNAUTHOR{Huang et al.} % for four or more authors
	% \RUNAUTHOR{Huang and Wang }
	
	\RUNTITLE{ORLM: A Customizable Framework in Training Large Models for Automated Optimization Modeling}
	% \TITLE{Bundling Information Goods of Decreasing Value}
	\TITLE{ORLM: A Customizable Framework in Training Large Models for Automated Optimization Modeling} %\footnote{The authors' names are alphabetically ordered. }}

\ARTICLEAUTHORS{\footnotesize
 	\AUTHOR{\textbf{ Chenyu Huang$^{1}$\footnotemark[1], Zhengyang Tang$^{2,3}$\footnotemark[1],  Shixi Hu$^5$, Ruoqing Jiang$^6$\\Xin Zheng$^7$,  Dongdong Ge$^4$\footnotemark[2], Benyou Wang$^{2,3}$, Zizhuo Wang$^{2}$\footnotemark[2]}}
 	\AFF{$^1$ Shanghai University of Finance and Economics\\$^2$ The Chinese University of Hong Kong, Shenzhen (CUHK-Shenzhen)\\ $^3$ Shenzhen Research Institute of Big Data\\ $^4$ Shanghai Jiao Tong University\\ $^5$ Cardinal Operations\\ $^6$ Columbia University\\$^7$ Duke University}
\AFF{\textit{chenyuhuang@stu.sufe.edu.cn\quad
zhengyangtang@link.cuhk.edu.cn\quad
shixi@shanshu.ai\quad
rj2556@columbia.edu\quad
\\
xin.zheng@duke.edu\quad
ddge@sjtu.edu.cn\quad
\{wangbenyou,wangzizhuo\}@cuhk.edu.cn}}
} 
\footnotetext[1]{$*$ These authors contributed equally to this work.}
\footnotetext[2]{$\dag$ Corresponding author.}

% should be entered in ONE field, separated by a comma. 
% \EMAIL field can be repeated if more than one author
%\ARTICLEAUTHORS{%
 	%\AUTHOR{Chenyu Huang}
 	%\AFF{}
 	%\EMAIL{}
% 	% % Enter all authors
%} 
% Enter the full title:\footnote{The authors' names are alphabetically ordered.}}

% should be entered in ONE field, separated by a comma.
% \EMAIL field can be repeated if more than one author

\ABSTRACT{Optimization modeling plays a critical role in the application of Operations Research (OR) tools to address real-world problems, yet they pose challenges and require extensive expertise from OR experts. With the advent of large language models (LLMs), new opportunities have emerged to streamline and automate such task. However, current research predominantly relies on closed-source LLMs such as GPT-4, along with extensive prompt engineering techniques. This reliance stems from the scarcity of high-quality training datasets for optimization modeling, resulting in elevated costs, prolonged processing times, and privacy concerns. To address these challenges, our work is the first to propose a viable path for training open-source LLMs that are capable of optimization modeling and developing solver codes, eventually leading to a superior ability for automating optimization modeling and solving. Particularly, we design the {\sc OR-Instruct}, a semi-automated data synthesis framework for optimization modeling that enables customizable enhancements for specific scenarios or model types. This work also introduces IndustryOR, the first industrial benchmark for evaluating LLMs in solving practical OR problems. We train several 7B-scale open-source LLMs using synthesized data (dubbed ORLMs), which exhibit significantly enhanced optimization modeling capabilities, achieving competitive performance across the NL4OPT, MAMO, and IndustryOR benchmarks. Additionally, our experiments highlight the potential of scaling law and reinforcement learning to further enhance the performance of ORLMs. The workflows and human-machine interaction paradigms of ORLMs in practical industrial applications are also discussed in the paper.
}

% \HISTORY{This paper was
% first submitted on April 12, 1922 and has been with the authors for
% 83 years for 65 revisions.}

\maketitle

\section{Introduction}
\label{introduction}

Optimization models have been a critical analytical tool for business decision-making, aiding companies in making optimal choices under various complex decision environments. With the fast development of computing capability (including both the development of algorithm and hardware) and the growing amount of data, optimization models are playing more and more significant roles in the operations of large corporations. For instance, optimization techniques used by JD.com enable them to handle 10 times the normal volume of orders during peak sales seasons, helping the company decrease its fulfillment expense ratio to a world-leading level of 6.5\%, which has resulted in hundreds of millions of dollars in savings \citep{qin2022jd}. Baosteel, one of the largest steel company in China, utilized integer optimization for production planning at its main plant in Shanghai, leading to a significant reduction in carbon monoxide emissions — over 500,000 tons annually — as reported by \cite{INFORMS2013}. The U.S. Census Bureau was recognized as a finalist for the 2022 Franz Edelman Award for their innovative use of operations research and analytics in reengineering field operations for the 2020 U.S. Census. This project utilizes advanced analytics, optimization, and machine learning techniques to automate and optimize the scheduling, workload assignments, and management of field data collection, a process previously done manually \citep{adams2023advanced}. Despite the huge amount of success stories, the sensitivity of optimization models to environmental changes necessitates frequent adjustments in dynamically evolving business contexts. This, combined with the high level of expertise and substantial labor costs required for optimization modeling, represents a significant bottleneck that has hindered the full potential to be unlocked for optimization techniques.

The advent of large language models (LLMs) gives an emerging hope to the above challenge. As soon as its advent, these pre-trained LLMs have exhibited remarkable capabilities in rapid knowledge comprehension and cross-disciplinary learning, demonstrating remarkable success and transformative potential across various fields \citep{huang2024beyond}, {like medical consultation \citep{wang2023huatuo}, software development \citep{OpenAICodex2021}, offering stock market trading strategies \citep{chen2022expected}, improving recommendation systems \citep{xu2024prompting}, and enhancing marketing research \citep{wang2024large}.
In the realm of more abstract mathematical reasoning, research has increasingly incorporated more techniques to enhance the capabilities of large models, yielding promising results. For example, \cite{trinh2024solving} train Alphaproof with hundreds of billions of synthetic data used to tackle complex olympiad geometry problems. Similarly, MathScale \citep{tang2024mathscale} offers a straightforward and scalable method to generate high-quality data for mathematical reasoning, establishing a foundation for further training of large language models in mathematical reasoning capabilities. }

\begin{figure}[!htb]
    \centering
    \includegraphics[width=0.9\linewidth]{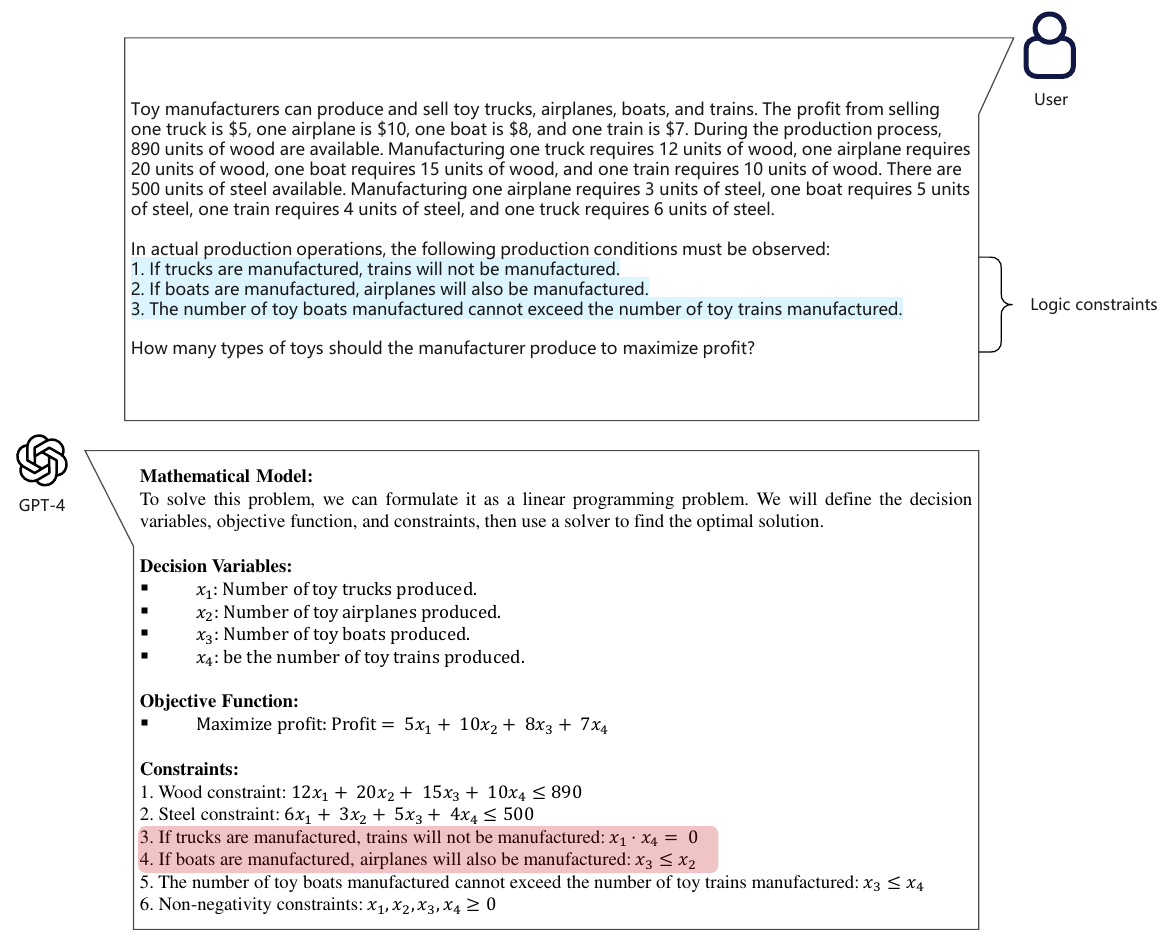}
    \caption{An example is presented to illustrate the failure of advanced large models to accurately respond to seemingly straightforward optimization modeling questions. The responses highlighted in reds contain errors, specifically the introduction of non-linear terms and misunderstandings of logical relationships. This response was generated in June 2024, originating from the web-based GPT-4.}
    \label{fig:chat}
\end{figure}

{In recent years, with the advancements in pre-trained language models (PLMs) such as GPT-4 and Claude-3.5, a growing number of studies have explored automating the modeling and solving of optimization problems expressed in natural language. To produce complete solutions, including executable programs that leverage established solvers, existing approaches often employ prompt engineering techniques and multi-agent collaboration frameworks built on proprietary LLMs, such as Chain-of-Experts~\citep{CoE}, OptiMUS~\citep{optimus}, and OptiGuide~\citep{optiguide}. However, utilizing online APIs for advanced large models offers no assurances of privacy protection, which is prohibitive in some sensitive industry applications. Additionally, these approaches are heavily reliant on the inherent capabilities of the foundational LLMs themselves, rendering them incapable of addressing the customized needs of users.} As illustrated in Figure \ref{fig:chat}, even a basic optimization modeling problem presents challenges for the cutting-edge foundational large language model, GPT-4. It demonstrates deficiencies in comprehending and transforming logical constraints, which imposes a limitation on methods dependent on these general closed-source LLMs.

In summary, existing large language models (LLMs) for optimization modeling exhibit the following deficiencies:
\begin{enumerate}
    \item \textbf{Limited modeling capabilities.} Contemporary advanced LLMs frequently exhibit errors in interpreting logical constraints or inadvertently incorporate nonlinear elements, making it difficult to model real-world problems effectively.
    \item \textbf{Insufficient quality of optimization modeling training data.} The quality and scale of data are directly proportional to the capability of the model, as described by the \textit{Scaling Law} \citep{Kaplan2020Scaling}. However, a shortage in high-quality optimization modeling training data is often a critical challenge, obstructing the advancement of LLMs' modeling ability.
    \item \textbf{Data privacy concerns.} Dependence on APIs for closed-source LLMs may lead to data breaches which hampers the broader adoption and development of LLM technologies in the industrial sector.
    \item \textbf{Test sets are relatively homogeneous.} The efficacy of large language models in automatic modeling is predominantly assessed using the dataset provided by the NL4OPT competition \citep{nl4opt}. This dataset primarily focuses on simple linear programming tasks that exhibit relatively low complexity, narrow scope, and few types compared to real-world applications.
\end{enumerate}

To fill this gap, this paper introduces, for the first time, a new path for training an open-source large language model specifically designed for modeling and solving optimization problems. To address the core of the above challenges, we present a semi-automated framework for the synthesis of high-quality optimization modeling data, aimed at progressively enhancing the specialized modeling capabilities of LLMs. We hope that this approach will inspire future research and serve as a foundation for the deeper integration of large models with operations research. Our main contributions can be summarized as follows:

\begin{enumerate}
    \item To ensure the effectiveness, robustness, and real-world applicability of our model, we have identified four crucial requirements that the training dataset must fulfill, as informed by academic research like \cite{alzubaidi2023survey,jiangPeekTokenBias2024} and industry insights.  Inspired by these requirements, we design and implement {\sc OR-Instruct}, a semi-automated process for creating synthetic data tailored to optimization modeling. The process uses an iterative bootstrapping algorithm (see Figure \ref{fig:OR-Instruct}). Initially, we collect a set of seed industry cases (e.g., 686 cases in our study) and add them to the training data pool. Following this, we use two strategies. {One is expansion, which employs GPT-4 and leverages its in-context learning capability to generate data spanning diverse scenarios and question types. Another is the augmentation strategy, which involves modifying objectives and constraints, rephrasing questions, and incorporating diverse modeling techniques to ensure the difficulty and quality of generation problems. 
    Finally, heuristics are used to automatically filter out obviously low-quality data. This cycle can be repeated multiple iterations until the training data pool reaches the requisite volume (e.g., 32,481 cases in our study). Notably, this approach is highly customizable; new seed data and scenarios can be seamlessly introduced to adapt the dataset to specific needs.}

\begin{figure}[!htbp]
\begin{center}
 \includegraphics[width=1.0\linewidth]{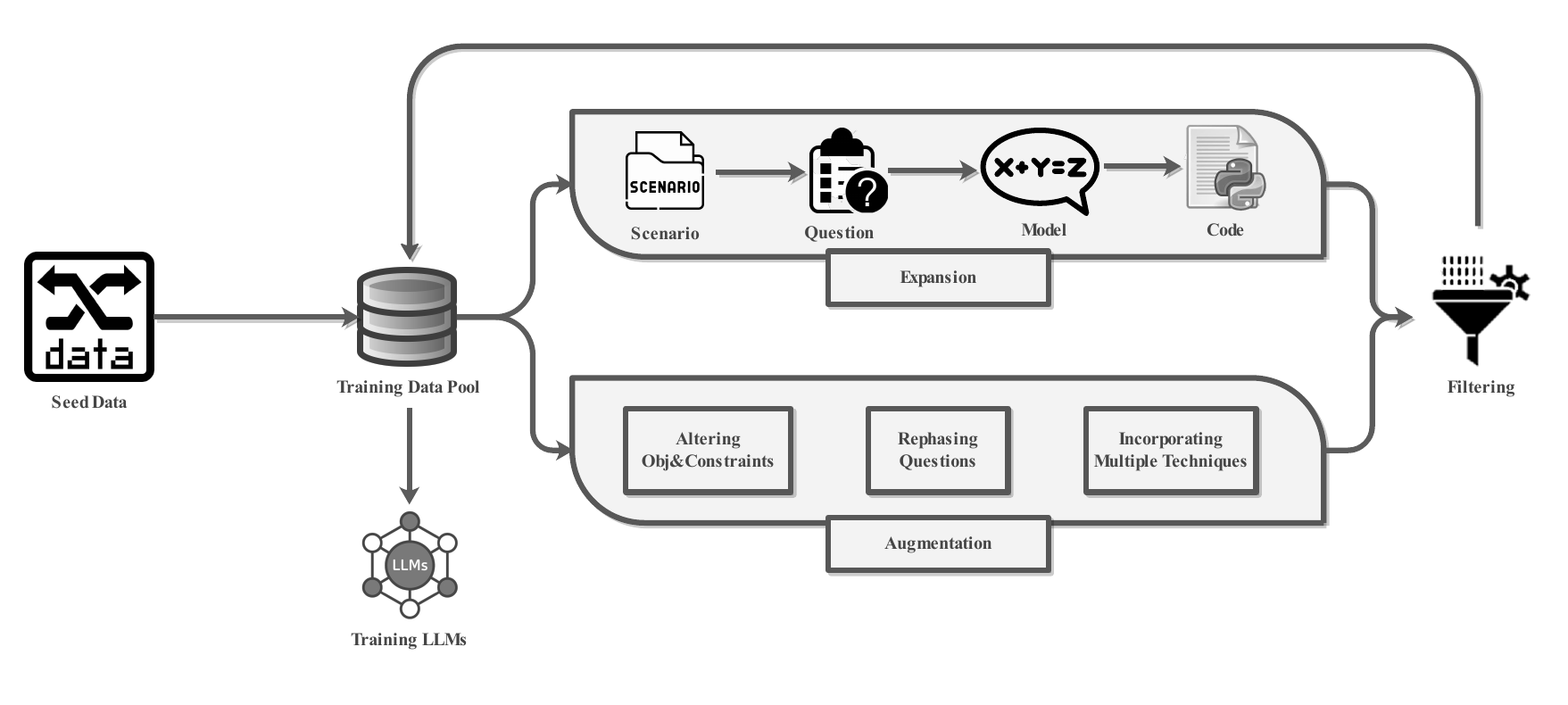}
\caption{Overview of {\sc OR-Instruct}.}
 \label{fig:OR-Instruct}
\end{center}
\end{figure}

\item To evaluate the efficacy of {\sc OR-Instruct}, we introduce IndustryOR, the inaugural industrial benchmark tailored for optimization modeling. This benchmark uses data sourced from 13 different industries, covering 5 types of questions across 3 levels of difficulty. For comprehensive evaluation, we also include NL4OPT~\citep{nl4opt} and MAMO~\citep{huang2024mamo} benchmarks.
\item We utilize the synthetic data generated by {\sc OR-Instruct} to train a series of open-source large language models, each approximately 7 billion parameters in size, including Mistral-7B~\citep{jiang2023mistral}, Deepseek-Math-7B-Base~\citep{deepseekmath}, and LLaMA-3-8B~\citep{llama3modelcard}. We designate these trained models as Operation Research Language Models (ORLMs) and observe a marked enhancement in their optimization modeling capabilities. Our experimental results show that the best-performing ORLM achieves competitive performance on the NL4OPT, MAMO, and IndustryOR benchmarks.  Building upon this foundation, we conduct extensive numerical experiments to analyze the limitations of the ORLM and propose feasible solutions for its improvement. The results highlight the promising prospects of ORLM in advancing the application and development of operations research, while also suggesting potential human-LLM collaboration frameworks.  These frameworks demonstrate how ORLM could enhance the productivity of both industry and academia in the field of operations research.
\end{enumerate}

To our best knowledge, this work is the first initiative to train open-source LLMs specifically for optimization modeling in real-world industries. The proposed approach significantly reduces the reliance on closed-source LLMs thus adding significant flexibility to the training process and greatly enhancing privacy preservation for industrial applications. Remarkably, the highest-performing ORLM achieves competitive results on the NL4OPT, MAMO, and IndustryOR benchmarks. This suggests that in the field of optimization modeling, our approach enables ORLMs with approximately 7 billion parameters to outperform many cutting-edge large language models, such as GPT-4, Llama-3.1, Qwen 2.5, and related agent-based frameworks.

We also conduct a thorough analysis of the ORLM's results, with numerous ablation experiments validating the effectiveness of our approach. The results show that, although ORLM performs relatively weak on some challenging datasets, several strategies can still ensure improvements in its performance. For instance, applying scaling laws to further increase the model size and data scale leads to continuous performance enhancements. Additionally, we find that ORLM has weak ranking capabilities for optimal optimization models. Using alignment strategies, such as reinforcement learning, also helps strengthen ORLM's performance.

Furthermore, we want to comment that the proposed data synthesis method, the \textit{OR-Instruct}, has the ability to facilitate customization to address particular needs of larger models, such as supply chain management, scheduling, inventory issues, or enhancing capabilities in linear programming, integer programming, and mixed integer programming. This synthetic data framework robustly addresses the scarcity of training data for optimization modeling, provides a valuable reference for advancing the application of operations research in various industrial sectors with the support of large models. In Section \ref{sec: Potential Applications}, we provide a detailed discussion of representative workflows driven by large models.

Finally, we note that almost all of our results, including certain samples from the training data set, industry-level benchmark for optimization modeling, and all the best-performing LLMs that have been fine-tuned, are made open source. We hope that our work can serve as a foundation for subsequent explorations, empowering future researchers to enhance the utility and efficiency of large language models in optimization modeling.

The structure of this paper is organized as follows: Section 2 reviews the literature related to our research. Section 3 delineates four essential conditions required for the training data and outlines the implementation specifics of the semi-automated synthetic data framework for optimized modeling, {\sc OR-Instruct}. Section 4 examines the impact of training open-source large models on various test sets and presents the experimental outcomes of customized enhancements in modeling capabilities. {Section 5 analyzes the limitations of ORLM through extensive experiments and discusses potential areas for improvement. Section 6 discusses the practical application scenarios of ORLM and explores future research directions based on the conclusions drawn in this paper.} We conclude the paper in Section 7.

\section{Literature Review}

Our research is related to several streams of literature, we will provide a detailed review of each stream below.

~\\
{\bf AI approach for facilitating solution to operations research problems.} With the fast development of AI techniques, there have been a growing amount of research leveraging advanced techniques, such as deep learning, to address challenging operational research (OR) problems traditionally resistant to conventional methods. For example, in the supply chain context, \cite{gijsbrechts2022can} employ deep reinforcement learning to enhance inventory management, demonstrating parity with leading heuristics and approximate dynamic programming approaches. \cite{qi2023practical} investigate a data-driven multi-period inventory replenishment problem with uncertain demand and vendor lead time, propose a one-step end-to-end framework that uses deep learning models to output the suggested replenishment amount directly from input features. {In the revenue management field, \cite{aouad2022representing} propose a neural network-based discrete choice model called RUMnets, which combines the theoretical robustness of traditional economic models with the flexibility and learning capabilities of modern artificial intelligence approaches, leading to better predictions and insights into consumer behavior.} \cite{wang2023neural} explore the deployment of deep neural networks in making the optimal assortment selection, demonstrating its effectiveness in capturing consumer choice. As for queuing theory, \cite{dai2022queueing} investigate the deployment of advanced policy gradient methods for controlling multi-class queueing networks, underscoring the superior efficacy of the PPO algorithm under diverse load scenarios. For personalized recommendation, \cite{wang2023tgvx}  utilize a novel unsupervised deep learning network and graph embedding techniques to improve recommendation accuracy and diversity by effectively transferring check-in patterns across different geographical contexts. Regarding general Markov decision process, \cite{zhang2021model} introduce a model-free reinforcement learning algorithm that attains asymptotically optimal sample complexity for learning an $\epsilon$-optimal policy within discounted Markov decision processes (MDPs).  In addition, these techniques have been used in demand prediction, for example, \cite{leeTransformerConformalPrediction2024} propose a Transformer-based methodology for time series conformal prediction, utilizing its decoder to estimate prediction intervals through the quantile prediction of residuals, showcasing enhanced performance over existing conformal prediction techniques. \cite{chen2020probabilistic} introduce a novel architecture of convolutional neural networks (CNNs) that significantly improves forecasting by modeling the conditional distribution of future series data directly, rather than relying on traditional autoregressive methods. 

{AI technology also demonstrates significant strengths in facilitating the implementation of operations research, for instance, \cite{liu2023deep} employ reinforcement learning combined with spatial data to optimize the allocation of resources and strategize for disease outbreak responses.  \cite{parmentier2022learning} introduces a novel paradigm that integrates machine learning with operations research to efficiently approximate complex industrial problems by transforming them into classic, more tractable optimization challenges.}

In addition, there have also been recent attempts in using AI methods to empower optimization algorithm design. For instance, deep learning or machine learning can be used to accelerate conventional optimization algorithms for tackling mixed-integer programming problems  (e.g., \citealt{chen2023pre,nair2020solving}). Numerous researchers also explore end-to-end optimization problem-solving strategies based on deep learning methods, motivated by the high structural similarity among many optimization problems, with differences primarily in data distribution (e.g., \citealt{chen2022learning,khalil2017learning}). Compared to this work, the above stream of works studies the integration of AI approach and traditional OR approaches in solving OR problems. In contrast, the main focus of the present work is in the facilitation and automation of the modeling process of the optimization model, which is different from the above stream.

~\\
\textbf{Automated optimization modeling} {is an emerging area, which aims to automate mathematical modeling and solving tasks to facilitate efficient decision-making. Recently, a growing body of studies have emerged, aiming to bridge the gap between natural language processing and mathematical optimization.} One of the pioneering endeavors in this field is the NL4OPT competition, which focuses on formulating mathematical models for optimization problems~\citep{nl4opt}. In particular, their work introduces a two-step framework using PLMs and offers a widely used benchmark for intelligence mathematical modeling. Expanding on this, \cite{CoE} introduce Chain-of-Experts, a multi-agent LLM framework to build optimization model that significantly outperforms GPT-4 by incorporating domain-specific knowledge and cooperative reasoning strategies. \cite{optimus} propose OptiMUS, a robust LLM-based agent designed to formulate and solve MILP problems. This system not only develops mathematical models but also assists in writing and debugging solver code, showcasing a comprehensive solution that extends beyond simple problem formulation. Similarly, \cite{optiguide} develop a framework called OptiGuide which leverages LLMs to interpret and explain optimization outcomes in supply chain management, thereby enhancing stakeholder comprehension and trust. In terms of diagnosing and resolving optimization issues, \cite{chenDiagnosingInfeasibleOptimization2023} introduce OptiChat, a natural language-based system equipped with GPT-4, which assists in identifying and addressing infeasible optimization models. {Unlike these methods, our approach involves fine-tuning open-source LLMs and delivering a complete solution using direct prompting, which is more concise and privacy-preserving than multi-agents interaction and simply calling API from advanced LLMs like GPT-4.}

~\\
\textbf{Synthetic data} {plays a crucial role in enhancing the capabilities of large language models by providing controlled, high-quality datasets that help fill gaps in real-world data, eliminate biases, and improve model robustness.} A review of literature on synthetic data and LLMs is referred to \cite{ding2024data}. {In the field of synthetic data for optimization modeling, traditional methods often require specialized knowledge to convert real-world problems into mathematical models, which limits their accessibility and flexibility. Most research focuses on how to automatically generate mixed-integer linear programming (MILP) models to overcome these challenges more effectively.}  For example, the methods proposed by \cite{pawlakAutomaticSynthesisConstraints2017} and \cite{srokaOneclassConstraintAcquisition2018} utilize MILP examples for the synthesis of constraints. These methods develop an approach that can generate constraints from examples of both feasible and infeasible solutions, thereby demonstrating the feasibility of automating this conversion process. Further expanding on this theme, \cite{pawlakGrammaticalEvolutionConstraint2021} leverage evolutionary strategies to generate constraints that improve the adaptability and efficiency of optimization models, thus enhancing the model’s flexibility in handling complex scenarios. \cite{prasath2023synthesis} examine the effectiveness of synthesizing mathematical programs from natural language specifications based on NL4OPT competition dataset, focusing on data synthesis methods using the CodeT5 model with enhancements from data augmentation and post-processing. The study highlights how synthetic data generation, notably through techniques like reverse translation and parameter variation, significantly improves model performance in solving optimization problems. {The synthetic data framework proposed in \cite{li2023synthesizing} leverages LLMs to identify decision variables, classify constraints, and generate MILP models based on NL4OPT problems.} This approach significantly outperforms traditional methods, achieving higher accuracy in constraint recognition and model formulation. In contrast to the aforementioned synthetic data approaches, the method proposed in this paper is designed for the general purpose of synthesizing high quality data for optimization modeling from unstructured natural language descriptions. This allows for the development of synthesis strategies that are not only tailored to specific problem but also capable of customized data augmentation.

\section{Semi-Automated Framework for Data Generation}
In this section, we present the formal definition of the optimization modeling task. We first discuss some critical requirements for effectively training an open-source LLM to perform this task. Based on these requirements, we design a semi-automated data synthesis framework {\sc OR-Instruct} which achieves the desired properties.

\subsection{Data Requirements}

\begin{figure*}[!htbp] \small
\begin{AIbox}{An Expected Training Example}
{\bf Input - Natural Language Problem:} \\
{\footnotesize
A company has three transportation options to choose from to transport $25$ tons of cargo, namely trucks, airplanes, and ships with costs $\$100, \$120, \$130$ per ton and capacities of $10, 20, 30$ tons respectively. The company can't choose trucks and ships together. How should the company optimize the selection and allocation of these methods to minimize overall costs?
}
\tcbline
{\bf Target - Mathematical Model and Program:} \\
{\footnotesize
\#\# Mathematical Model: \\
\#\#\# Variables: \\
- $x_1, x_2, x_3$: 0-1 variables indicating whether trucks, airplanes, and ships are selected, respectively. \\
- $y_1, y_2, y_3$: Non-negative continuous variables indicating the volume of cargo transported by three modes.\\
\#\#\# Objectives and Constraints:
\begin{align*}
\operatorname{Minimize}_{x_i\in \{0,1\},y_i\geq0} & \quad 100y_1 + 120y_2 + 80y_3 \\
\text{subject to} & \\
& x_1 + x_2 + x_3 \geq 1 & \text{(At least one transportation)} \\
& y_1 \leq 10x_1, \quad y_2 \leq 20x_2, \quad y_3 \leq 30x_3 & \text{(Capacity limits)} \\
& x_1 + x_3 \leq 1 & \text{(Mutual exclusion constraint)} \\
& y_1 + y_2 + y_3 \geq 25 & \text{(Transportation requirements)}
\end{align*}
\#\# Program using COPT solver:
\begin{lstlisting}[language=Python]
import coptpy as cp
from coptpy import COPT
# Initialize the optimization environment
env = cp.Envr()
model = env.createModel("TransportationOptimization")
# Define costs and capacities for each mode of transportation
costs = {'trucks': 100, 'airplanes': 120, 'ships': 130}
capacities = {'trucks': 10, 'airplanes': 20, 'ships': 30}
# Define binary and continuous variables for each transportation mode
x = {mode: model.addVar(vtype=COPT.BINARY, name=f"x_{mode}") for mode in costs}
y = {mode: model.addVar(vtype=COPT.CONTINUOUS, lb=0, name=f"y_{mode}") for mode in costs}
# Set the objective function to minimize the total transportation cost
model.setObjective(sum(costs[mode] * y[mode] for mode in costs), sense=COPT.MINIMIZE)
# Add constraints
model.addConstr(x['trucks'] + x['airplanes'] + x['ships'] >= 1, name="AtLeastOneMode") 
for mode in costs:
    model.addConstr(y[mode] <= capacities[mode] * x[mode], name=f"Capacity_{mode}")  
model.addConstr(x['trucks'] + x['ships'] <= 1, name="ModeExclusivity")  
model.addConstr(sum(y[mode] for mode in capacities) >= 25, name="Volume Requirement")  
# Solve the model
model.solve()
# Check the solution status and print the optimal values of the variables
if model.status == COPT.OPTIMAL:
    print("Optimal solution found:")
    for mode in costs:
        print(f"{mode}: x = {x[mode].x}, y = {y[mode].x}")
\end{lstlisting}
}
\end{AIbox}
\caption{An expected training example for optimization modeling task.}
\vspace{-15pt}
\label{fig:training_example}
\end{figure*}

%\subsection{Definition of Optimization Modeling}
Given an  OR problem \(p\) described in natural language, optimization modeling \citep{berry1995mathematical,optimus} involves constructing a mathematical model \(m\) that converts the real-world problem into an optimization model with concrete decision variables, objective function and constraints. Then, for large-scale problems, in order to obtain an optimal solution, we further need to convert the mathematical model into a program $c$ compatible with a specific optimization solver. Hence, an expected training example for this task is usually required in the form of the triplet \((p, m, c)\), as illustrated in Figure \ref{fig:training_example}. Training a large language model \(f\) for this task essentially involves learning a function \(f: p \rightarrow (m, c)\) that maps problems to their corresponding mathematical models and solver programs. In this paper, without loss of generality, COPT is employed as the default solver. COPT~\citep{copt}, an acronym for Cardinal Optimizer, is able to tackle large-scale optimization problems and achieves good performance in a variety of tasks \citep{mittelmann2002benchmark}. We select the COPT solver because the predominant large language models available in the market have not yet mastered the COPT syntax. This choice underscores the efficiency and extensibility of our framework. However, any well-established solver may be utilized in the training dataset construction.

%\subsection{\emph{Desiderata} of Training LLMs for Optimization Modeling}
%\label{sec:Desiderata}

In addition to the desired format as described above, a desired training dataset should demonstrate effectiveness, robustness, and practical applicability to real-world scenarios. In particular,
 
\begin{proper}\label{Coverage}
\textbf{Comprehensive Coverage:} 
The dataset should cover:
1) diverse scenarios such as supply chain optimization, scheduling, inventory management, and transportation logistics;
2) different problem types like linear programming, integer programming, and mixed integer programming;
and 3) varying difficulty levels (easy, medium, hard) --- we provide a way to describe the difficulty level of a problem in Appendix \ref{app:criteria_of_difficulty}.
\end{proper}

\begin{proper}\label{Adaptability}
\textbf{Environmental Adaptability:} In real-world industrial settings, the objectives and constraints of problems often change due to shifts in business goals, market conditions, or resource availability. A good optimization model should be able to adapt to those changes easily. Therefore, it is vital that the dataset includes cases reflecting these dynamic changes.
\end{proper}

\begin{proper}\label{Diverse}
\textbf{Linguistic Diversity}:  Problems described in natural language often show different syntax, ambiguities, and complexities. For example, one problem might mention "inventory overflow" while another refers to "excess stock". Including this Linguistic Diversity in the dataset could improve the model's ability to understand varied descriptions.
\end{proper}

\begin{proper}\label{Variability}
\textbf{Technique Variability:} For some challenging problems, there may be multiple modeling techniques, such as linearizing a nonlinear problem by introducing auxiliary variables, or introducing the big $M$ term in an integer optimization formulation. Including this variety in the dataset allows the model to learn different modeling techniques and approaches.
\end{proper}

\subsection{{\sc OR-Instruct}: Towards Training Effective OR LLMs}
\label{sec:or_instruct}

In the above section, we describe some critical requirements on the dataset for an effective training of an LLM to be equipped with good optimization modeling capabilities. However, collecting data on a large scale that aligns with the above requirements presents significant challenges because: (1) such data primarily resides within private industrial cases, and no public datasets sufficiently meet the requirements, (2) training data suitable for optimization modeling is limited and requires extensive time to collect, and (3) conventional synthesis methods (for example,~\citealt{selfinstruct}) encounter substantial obstacles in generating this type of data, as will be elaborated below. In the following, to address the above challenges, we design and implement {\sc OR-Instruct}, a semi-automated process for creating synthetic data tailored to these requirements. The overall pipeline of OR-Instruct is depicted in Figure \ref{fig:OR-Instruct}, which iteratively applies two strategies, namely expansion and augmentation, to the training data pool, followed by the filtering of obviously low-quality data.

\subsubsection*{Initial Dataset}
\label{sec:Initial Dataset}

{ The quality of initial data plays a crucial role in large model training, providing a strong foundation for diversity and difficulty as a starting point \citep{marion2023less}. Additionally, in this customizable framework designed to enhance modeling capabilities, the category or domain of the optimization problem associated with the initial data is important. We will demonstrate later that when the initial dataset is concentrated in a specific type (e.g., scenario or type), it substantially enhances the LLM's ability to model effectively within that domain.

In our particular case, we start with 686 real-world industry cases collected from some OR textbooks and our previous industrial projects (after proper abstraction and anonymization). }

\subsubsection*{Strategy 1: Expansion for Comprehensive Coverage.}
\label{sec:expansion}
{ Initially, {\sc OR-Instruct} seeks to generate new data by expanding scenarios and question types from the training dataset using a bootstrapping approach powered by GPT-4. The process begins with GPT-4 generating a list of 100 scenarios where optimization models are applied in real-life contexts. In each iteration, three examples are selected from data pool to serve as in-context references for GPT-4, and one scenario is randomly chosen from the list. Among the three examples, two are sourced from real-world data, while the third, if available, is taken from data previously generated by the model. Finally, GPT-4 is prompted to create a new sample based on the three examples within the context of the selected scenario. This iterative approach promotes greater diversity in the resulting dataset. The prompting template for this expansion process is detailed in Appendix \ref{app:prompt_for_expansion}.

It's important to note that optimization models are naturally abstract, often built on similar logical ideas across various scenarios. In this setting, \textit{Comprehensive Coverage} not only exposes the LLMs to a broad range of situations but also helps it better grasp the underlying patterns and principles when modeling across these different contexts.} However, it falls short of meeting the other requirements, especially concerning varying levels of difficulty. In manually reviewing the difficulty of generated example, of 50 cases, 87\% are deemed easy, 13\% medium, and none hard, as judged by criteria in Appendix \ref{app:criteria_of_difficulty}. We also provide examples in Appendix \ref{app:compare_easy_and_hard} for comparing generated easy entries with real-world hard entries. Fortunately, the original seed data already shows a diverse range of difficulties. Thus, we can naturally enhance the difficulty diversity by augmenting them described in the next strategy, thereby further addressing \textit{Comprehensive Coverage}.

\subsubsection*{Strategy 2: Augmentation for Problem-Solution Diversity.}

To fulfill the remaining requirements, the {\sc OR-Instruct} adopts an augmentation strategy for the original seed data, designed to both enhance the diversity and robustness of the dataset while preserving the complexity inherent in optimization modeling problems. This strategy is crafted to mirror potential scenarios that might be encountered in real-world industrial applications, encompassing modifications such as alterations in problem descriptions, model adjustments, or concurrent changes. To address these three scenarios, the implementation of this strategy involves three specific tactics: rephrasing questions, modifying objectives and constraints, and integrating diverse modeling techniques. Together, these tactics aim to broaden the range of problem-solution diversity, thereby enriching the training dataset.

\noindent
\begin{enumerate}
    \item[(a)] \textbf{Altering Objectives and Constraints:} The first augmentation involves adding, removing, or replacing objectives and constraints in the problem, along with making necessary adjustments to the mathematical models and programs. Specifically, we start by providing GPT-4 with the original example and ask it to list five potential changes to the objectives and constraints. These suggested changes are then fed back to GPT-4 using a prepared few-shot prompt to modify the problem, model, and programs accordingly. This augmentation is designed to enhance \textit{Environmental Adaptability}.   An example is provided in Figure \ref{fig:adaptability_example}, and the corresponding prompting template is shown in Appendix \ref{app:prompt_for_altering}. 
    
    \begin{remark}
        {One important point to note is that during the process of altering objectives and constraints, GPT-4 may generate background scenarios that are illogical, such as presenting two contradictory constraints. However, this issue arises from unreasonable problem descriptions. As long as the model can accurately map each statement of the problem to the correct objectives and constraints, it can still yield the desired results when tested on meticulously designed datasets that reflect such issues.}
    \end{remark}

\begin{figure*}[!htbp] \small
\small
\begin{AIbox}{Altering Objectives and Constraints for Requirement \textcolor{white}{\ref{Adaptability}}}
% {\bf Altering Objectives and Constraints} 
% \tcbline
{\bf Original:} \\
{\footnotesize
Q: ... The company can't choose trucks and ships together. Denote the cost ...
}
\tcbline
{\bf Augmented:} \\
{\footnotesize
Q: ... The company can't choose trucks and ships together. \textit{Due to the special nature of the goods, the company has decided that if trucks are chosen, airplanes must also be selected for transportation.} Denote the cost ...

A: ...  
\textit{New dependency constraint (choosing trucks necessitates choosing airplanes): $x_1 \leq x_2$}
...
\begin{lstlisting}[language=Python]
...
model.addConstr(x['trucks'] <= x['airplanes'], name="New constraint")
...
\end{lstlisting}
}
\end{AIbox}
\caption{An example illustrating altering objectives and constraints. \textit{Italicized text} denotes altered content.}

\label{fig:adaptability_example}
\end{figure*}

\item[(b)] \textbf{Rephrasing Questions:} {The second augmentation involves modifying the problem description while essentially addressing the same optimization model. This approach enhances the robustness of large language models (LLMs) to different prompts, enabling them to abstract a unified optimization model across various problem descriptions.} This process involves instructing GPT-4 to rewrite the target problem, either simplifying or complicating it, while ensuring the core logic aligns with the solution, including the mathematical model and programs. This augmentation is designed to enhance \textit{Linguistic Diversity}. An example is provided in Figure \ref{fig:diverse_example}, and the corresponding prompting template is shown in Appendix \ref{app:prompt_for_rephrasing}.

\begin{figure*}[!htbp] \small
\begin{AIbox}{Rephrasing Questions for Requirement~\ref{Diverse}}
% {\bf Rephrasing Questions}
% \tcbline
{\bf Original:} \\
{\footnotesize
Q: A company has three transportation options to choose from to transport $25$ tons of cargo, namely trucks, airplanes, and ships with costs $\$100, \$120, \$130$ per ton and capacities of $10, 20, 30$ tons respectively. The company can't choose trucks and ships together. How should the company optimize the selection and allocation of these methods to minimize overall costs?
}
\tcbline
{\bf Augmented:} \\
{\footnotesize
Q: \textit{A corporation wants to transport 25 tons of cargo with least cost, and must choose from three transportation modes: trucks, airplanes, and ships. These options cost \$100, \$120, and \$130 per ton, respectively, with capacities of 10, 20, and 30 tons. However, trucks and ships cannot be used together.}
}
\end{AIbox}
\caption{An example illustrating rephrasing questions. \textit{Italicized text} denotes the question that has been rephrased.}

\label{fig:diverse_example}
\end{figure*}
\item[(c)] \textbf{Incorporating Multiple Modeling Techniques:} The third augmentation explores the use of different modeling techniques. We identify five potential techniques from engineers' experiences, such as introducing auxiliary variables or using the Big $M$ method, for GPT-4 to choose from in modifying an objective or constraint in the original mathematical model. This augmentation is designed to enhance \textit{Solution Variability}. An example is provided in Figure \ref{fig:variability_example}, and the corresponding prompting template is shown in Appendix \ref{app:prompt_for_multiple}.

\begin{figure*}[!htbp] \small
\begin{AIbox}{Incorporating Multiple Modeling Techniques for Requirement~\ref{Variability}}
% {\bf }
{\bf Original:} \\
{\footnotesize
A: Mutual exclusion constraint (trucks and ships cannot be selected simultaneously): $x_1+x_3 \leq 1$
}
\tcbline
{\bf Augmented:} \\
{\footnotesize
A: \textit{Mutual exclusion constraint (Using big $M$ method): $x_1 \leq (1-x_3)M, \text{where } M \text{ is a large number}$}
\begin{lstlisting}[language=Python]
...
model.addConstr(x['trucks'] <= (1-x['ships'])*M, name="New constraint")
...
\end{lstlisting}
}
\end{AIbox}
\caption{An example illustrating incorporating multiple modeling techniques. \textit{Italicized text} denotes the section that have been re-modeled using new techniques.}

\label{fig:variability_example}
\end{figure*}
\end{enumerate}

\subsubsection*{Postprocessing and Filtering}
\label{sec:filtering}
{ At the end of each iteration, {\sc OR-Instruct} implements correction and filtering measures on the generated examples. First, we apply a regular match correction function to address minor grammatical errors in the programs, which may arise from GPT-4's limited familiarity with the COPT API. Unexecutable programs are discarded, as they clearly represent low-quality data. Notably, in the final iteration, we also eliminate duplicate questions and any examples overlapping with the evaluation benchmarks to ensure the dataset remains uncontaminated.

Based on manual sampling assessments, we find that the accuracy of the synthesized data by expansion operation is about 70\% and 75\% for the augmentation data, judged by the correctness of both the code and the model. On average, each iteration filters out approximately 39\% of the generated examples, with the remaining examples incorporated into our training data pool.}

\vspace{0.4cm}

{In summary, the above outlines all the key steps of OR-Instruct, and the specific procedures can be found in Figure \ref{fig:OR-Instruct}.}

\section{Results}
\label{sec:experiments}

In this section, we train the open-source 7b size large language models with synthetic data from {\sc OR-Instruct} and compare the results with GPT-4 as well as other benchmarks. Then, using the case of mixed integer linear programming as an example, we demonstrate the customized enhancement effect of {\sc OR-Instruct} by including directionally generated training data.

\subsection{Data Generation by Using {\sc OR-Instruct}}

In our experiment, we initiate the study with 686 cases derived from previous industry projects, which the team engages with previously. These cases are abstracted and anonymized appropriately. Throughout the {\sc OR-Instruct} process, we utilize the proprietary LLM {\tt gpt-4-0613} as the standard model. For each cycle, the training data pool undergoes an expansion procedure 20,000 times and each augmentation operation is applied 6,000 times. Subsequently, an automatic filtration system excludes low-quality entries. After completing two iterations of this framework, we amass a total of 32,481 training examples.

\begin{figure}[!ht]    
\centering            
\subfloat[Distribution of industies]{
	\includegraphics[width=0.4\textwidth]{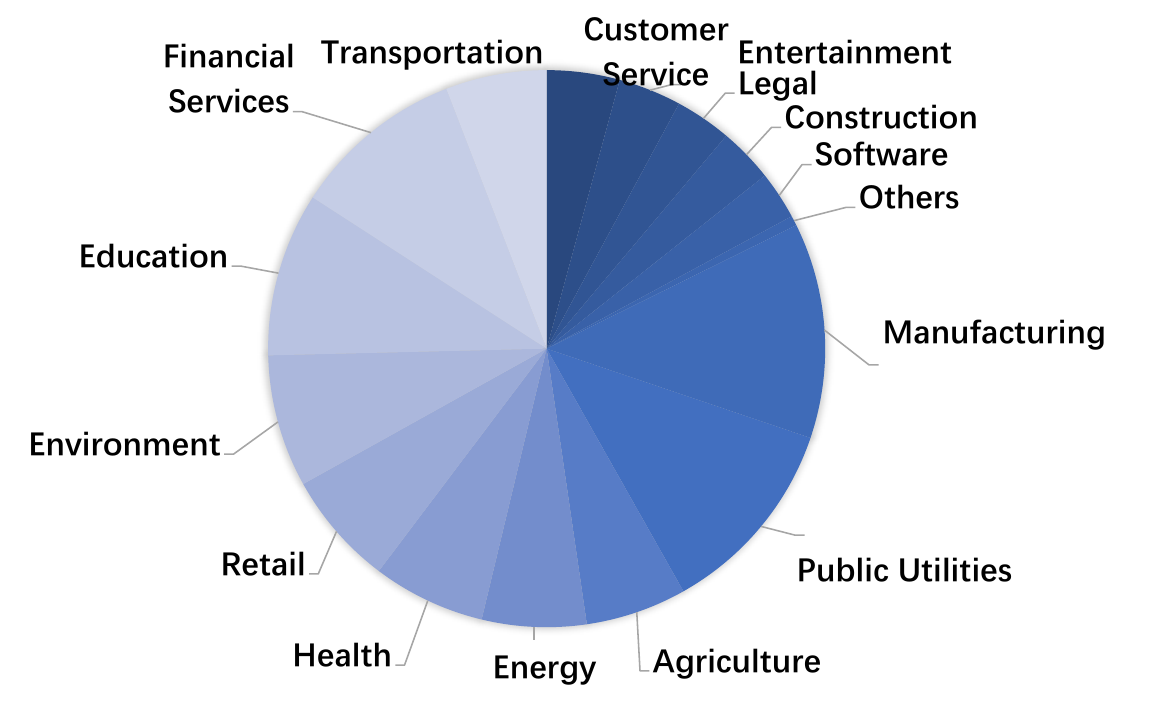}
}
\hspace{-2mm}
\subfloat[Question type]{
\vspace{0.4cm}
\includegraphics[width=0.5\textwidth]{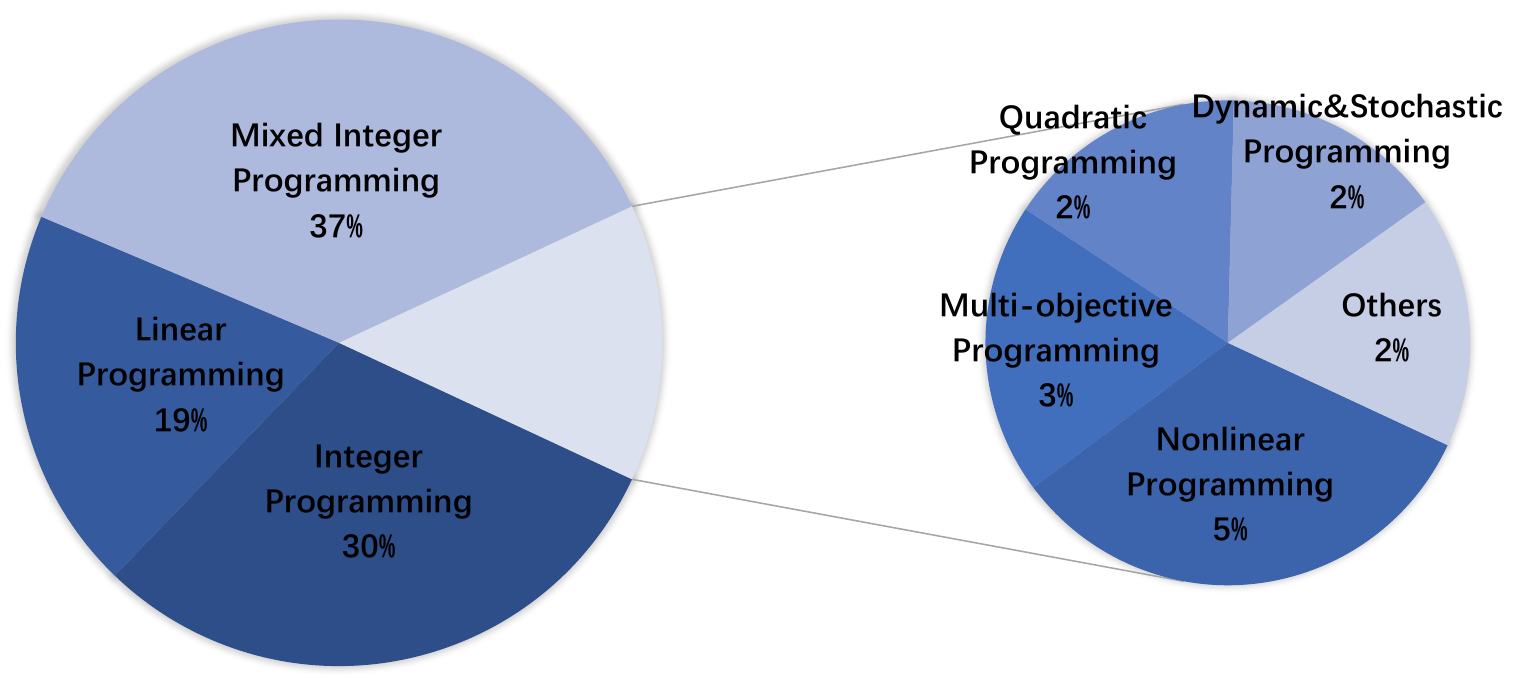}
}
\caption{Statistics of the data generated by {\sc OR-Instruct}}
\label{fig:stats}
\end{figure}

\noindent
\textbf{Statistics:} Figure \ref{fig:stats} presents the statistics of the {\sc OR-Instruct} Data. Of the data, 57\% is generated by the expansion operation, 17.2\% by the augmentation of altering objectives and constraints, 15.3\% by the augmentation of rephrasing questions, and 10.5\% by the augmentation of incorporating multiple modeling techniques. For scenario diversity, we have abstracted 16 industries from a total of 1,556 expanded scenarios to facilitate their representation in a plot. Regarding question type diversity, it expands to 8 question types. In terms of data quality, the accuracy of correctness are 70\% for expansion data and 75\% for augmentation data. We find these accuracy acceptable in exchange for automatic filtering and more cost-effective data synthesis. Additionally, our preliminary experiments have demonstrated the benefits of incorporating such data.

\subsection{Model Finetuning and Inference}

{ This paper employs instruction tuning for open-source large models, a technique designed to enhance the ability of large language models (LLMs) to follow diverse natural language instructions. The primary aim is to improve zero-shot performance on unseen tasks, facilitating effective generalization across various instruction-based scenarios. The methodology involves training on a dataset of instruction-output pairs, typically consisting of three elements: an instruction (e.g., "Translate this sentence from English to Spanish"), optional context, and the target output. The training can be mathematically represented as:
\[
\mathcal{L}(\theta) = -\sum_{(x, y) \in \mathcal{D}} \log P(y \mid x; \theta)
\]
where \(\mathcal{L}(\theta)\) is the loss function, \(x\) is the instruction input, \(y\) is the expected output, and \(\theta\) denotes the model parameters.} We borrow a widely used training framework from open-instruct~(\citealt{tulu}), where we take a natural language OR problem wrapped in an Alpaca-like template~(\citealt{alpaca}.) as the input prompt (see Appendix \ref{app:alpaca_template}), and treat a complete solution including mathematical models and programs as the target completion.

During training, we compute the loss only over the target completion. We employ the {\sc OR-Instruct} dataset to train several open-source large language models (LLMs) of approximately 7 billion parameters, including Mistral-7B~(\citealt{jiang2023mistral}), Deepseek-Math-7B-Base~(\citealt{deepseekmath}), and LLaMA-3-8B~(\citealt{llama3modelcard}), which serve as our backbone models.  { We conduct a grid hyper-parameter search for each backbone and present the best-performing hyperparameters in Table \ref{tab:ablation0}. }\footnote{Prior to the final fine-tuning, we conducted a grid search with learning rates within \{2e-5, 5e-5, 7e-5, 3e-6, 5e-6, 7e-6\}, batch sizes within \{64, 128, 256, 512\}, and training epochs within \{1, 2, 3\} to determine the optimal settings.} The models resulting from this process are referred to as ORLMs.

{Regarding the inference techniques during evaluation, it is important to consider that large language models, as probabilistic generative models, exhibit inherent randomness. This randomness can lead to inconsistent outcomes across multiple evaluations. To mitigate the influence of this randomness, we implemented a greedy decoding strategy in a zero-shot scenario, consistently selecting the highest probability output (top-1) as the final result. After obtaining these results, we executed the corresponding code to derive the predicted optimal values for comparison with the ground truth. Although this greedy approach may reduce the performance of large models on such tasks, it ensures the stability of the output and enables the reproducibility of our findings under identical conditions. We will discuss the impact of employing additional inference techniques on the performance of large models in Section \ref{sec: Inference Techniques}.}

\begin{table}[h]
\centering
\caption{Hyper-parameters for training ORLMs.}
\label{tab:ablation0}
\small{\begin{tabular}{lccc}
\toprule
\textbf{Backbone} & \textbf{BatchSize} & \textbf{LearningRate} & \textbf{Epochs}\\
\midrule
\makecell[l]{Mistral-7B}   & 512 & 3e-6 & 2\\ 
\makecell[l]{Deepseek-Math-7B-Base}   &128 & 2e-5& 2\\ 
\makecell[l]{LLaMA-3-8B}   & 64& 5e-6&2\\ 
\makecell[l]{Qwen-2.5-7B}   & 64& 2e-5&2\\ 
\bottomrule
\end{tabular}}
\end{table}

\subsection{Evaluation and Baselines}

\noindent
\textbf{Evaluation Benchmarks and Metrics:} We use NL4OPT~(\citealt{nl4opt}.), MAMO~(\citealt{huang2024mamo}.), and IndustryOR as evaluation benchmarks. NL4OPT is the most widely used benchmark for operations research and includes 289 linear programming problems in its test set. However, NL4OPT only provides mathematical models as targets, which complicates the verification of execution accuracy due to the absence of optimal solutions. To address this, we convert these mathematical models into programs using GPT-4, calculate and check the optimal solutions, and use these as ground truth. MAMO, a concurrent project, evaluates the mathematical modeling capabilities in LLMs. It includes 652 easy and 211 complex linear programming problems, each paired with its corresponding optimal solution, sourced from various academic materials. IndustryOR, which is proposed in this work, is the first industrial benchmark, consisting of 100 real-world OR problems from eight industries. It covers 5 types of questions,  linear programming, integer programming, mixed integer programming, non-linear programming, and others across 3 levels of difficulty. We measure performance using execution accuracy, where an executed optimal value that matches any provided ground truth optimal value is considered correct. Compared to NL4OPT, this metric enables a fully automated evaluation and provides greater flexibility for mathematical modeling approaches.

\noindent
\textbf{Baselines:} To ensure a comprehensive evaluation, we select a diverse set of models from previous methods for comparison. We include tag-BART~\citep{tagbart}, which secured the first place in the NeurIPS competition~\citep{ramamonjison2022augmenting}. Additionally, we consider methods that utilize proprietary LLMs. The "standard" prompting method involves prompting a proprietary LLM to produce mathematical programs, serving as a fundamental baseline. We also incorporate complex prompt engineering methods such as Reflexion~\citep{shinn2023reflexion}, Chain-of-Experts~\citep{CoE}, and OptiMUS~\citep{optimus}. These methods employ agents to refine both mathematical models and programs, achieving good performances on NL4OPT. We report their performance based on GPT-3.5 and GPT-4, respectively. Note that we implement the standard prompting on IndustryOR using the toolkit released by Chain-of-Experts~\citep{CoE}.

In addition, we introduce the most well-known open-source large models for comparison, including Llama-3.1-lnstruct \citep{llama3}, DeepSeek-v2-Chat \citep{deepseekv2}, DeepSeek-R1 \citep{guo2025deepseek}, Qwen2-Instruct \citep{qwen2}, and Mistral-Nemo \citep{mistralnemo}. Each model is tasked with constructing optimization models and generating code to solve the test set problems. 

We also assess the difference between the performance of ORLM and human-level performance. Given that constructing and solving optimization models requires a solid foundation in operations research and programming skills, we recruit 8 highly capable senior undergraduate students from relevant fields and 8 experts in the domain to evaluate four types of test sets. Each test set is evaluated by 2 senior undergraduate students and 2 experts. Each participant randomly selects 70 questions from their corresponding test set and is required to construct an optimized model and calculate the optimal solution. Participants are allowed to use any programming language; however, to ensure fairness in comparison with large models, they are only permitted to submit a finalized version of their code for testing and cannot make further modifications. Finally, the accuracy rates for the student and expert groups on each test set are obtained by averaging the results.

\begin{table}[!htbp]
\centering
\caption{Comparison of performance on the NL4OPT, MAMO, and IndustryOR benchmarks. Values marked with a \textsuperscript{*} are directly copied from original papers, with blanks where data were not reported. The highest results are highlighted in bold.}
\label{table:main}
\small{\begin{tabular}{lc|cccc|ccc}
\toprule
\textbf{Method/Model} & \textbf{Size} & \textbf{NL4OPT} & \makecell[c]{\textbf{MAMO}\\\textbf{EasyLP}} & \makecell[c]{\textbf{MAMO}\\\textbf{ComplexLP}} & \textbf{IndustryOR} & \makecell[c]{\textbf{Micro}\\\textbf{Avg}} & \makecell[c]{\textbf{Macro}\\\textbf{Avg}}\\
\midrule
\rowcolor{gray!17} \multicolumn{8}{c}{\textit{Methods based on PLMs}} \\
\makecell[l]{tag-BART} & 140/400M & 47.9\%\textsuperscript{*} & - & - & - & - & -  \\ 
\midrule
\rowcolor{gray!17} \multicolumn{8}{c}{\textit{Methods based on GPT-3.5}} \\
\makecell[l]{Standard} & {\footnotesize Unknown} & 42.4\%\textsuperscript{*} & - & - & - & - & -  \\ 
\makecell[l]{Reflexion} & {\footnotesize Unknown} & 50.7\%\textsuperscript{*} & - & - & - & - & -  \\ 
\makecell[l]{Chain-of-Experts} & {\footnotesize Unknown} & 58.9\%\textsuperscript{*} & - & - & - & - & -  \\ 
\midrule
\rowcolor{gray!17} \multicolumn{8}{c}{\textit{Methods based on GPT-4}} \\
\makecell[l]{Standard} & {\footnotesize Unknown} & 47.3\%\textsuperscript{*} & 66.5\%\textsuperscript{*} & 14.6\%\textsuperscript{*} & 28.0\% & 50.2\% & 39.1\%  \\   
\makecell[l]{Reflexion} & {\footnotesize Unknown} & 53.0\%\textsuperscript{*} & - & - & - & - & -  \\ 
\makecell[l]{Chain-of-Experts} & {\footnotesize Unknown} & 64.2\%\textsuperscript{*} & - & - & - & - & -  \\ 
\makecell[l]{OptiMUS} & {\footnotesize Unknown} & 78.8\%\textsuperscript{*} & - & - & - & - & -  \\ 
\midrule
\rowcolor{gray!17} \multicolumn{8}{c}{\textit{Standard prompting based on open-source LLMs}} \\
\makecell[l]{Llama-\scriptsize{3.1-Instruct}} & 405B & 38.7\% & 35.1\% & 20.8\% & 13.0\% & 31.5\% & 26.9\% \\
\makecell[l]{DeepSeek-\scriptsize{V2-Chat}} & 236B & 66.5\% & 60.5\% & 32.7\% & 16.0\% & 53.1\% & 43.9\% \\
\makecell[l]{Qwen2-\scriptsize{Instruct}} & 72B & 72.6\% & 79.9\% & 29.0\% & 18.0\% & 64.4\% & 49.8\% \\
\makecell[l]{DeepSeek-\scriptsize{R1-Distill}} & 32B & 80.4\% & 69.1\% & \textbf{45.4\%} & 33.0\% & 64.8\% & 56.9\% \\
\makecell[l]{Mistral-\scriptsize{Nemo}} & 12B & 14.6\% & 19.4\% & 3.7\% & 7.0\% & 14.6\% & 11.1\% \\
\midrule
\rowcolor{gray!17} \multicolumn{8}{c}{\textit{ORLMs based on open-source LLMs}} \\
\makecell[l]{ORLM-\scriptsize{Mistral}} & 7B & 84.4\% & 81.4\% & 32.0\% & 27.0\% & 68.8\% & 56.2\% \\
\makecell[l]{ORLM-\scriptsize{Deepseek-Math}} & 7B & \textbf{86.5\%} & 82.2\% & 37.9\% & 33.0\% & 71.2\% & 59.9\%  \\
\makecell[l]{ORLM-\scriptsize{LLaMA-3}} & 8B & 85.7\% & {82.3\%} & 37.4\% & \textbf{38.0\%} &  {71.4\%} & \textbf{60.8\%} \\ 
\makecell[l]{ORLM-\scriptsize{Qwen2.5}} & 7B & 86.1\% & \textbf{85.2\%} & 44.1\% & 25\% &  \textbf{73.7\%} & 60.1\%\\
\midrule
\rowcolor{gray!17} \multicolumn{8}{c}{\textit{Human Evaluation}} \\
\makecell[l]{Senior Undergraduates} & - & 80.4\% & 84.9\% & 53.1\% & 44.0\% & 75.2\% & 65.6\% \\
\makecell[l]{Experts} & - & 94.3\% & 90.4\% & 78.9\% & 76.0\% & 85.0\% & 88.2\%  \\
\bottomrule
\end{tabular}}
\end{table}

The results are presented in Table~\ref{table:main}. First, it is clear that methods based on LLMs generally outperform the PLM-based best method (tag-BART) in the NL4OPT test. This suggests that PLMs have limited generalization capabilities. For proprietary LLMs, as the mathematical reasoning capability increases from GPT-3.5 to GPT-4, we observe that performance has advanced across all prompt engineering methods. %Finally, ORLMs based on various open-source LLMs have demonstrated significantly improved optimization modeling capabilities, compared to vanilla open-source LLMs, which score 0 for all benchmarks in our preliminary experiments due to failing to output executable programs. This underscores the effectiveness of the {\sc OR-Instruct} data. 

Our best-performing ORLMs, like ORLM-{\scriptsize LLaMA-3} and ORLM-{\scriptsize Qwen2.5}, achieve state-of-the-art performance across four benchmarks with respect to GPT-4 and other open-source LLMs\footnote{Due to the severe scarcity of API resources, we are unable to conduct large-scale modeling tests on many state-of-the-art LLMs, such as DeepSeek-R1 (671B).}, surpassing Standard prompting based on GPT-4 by 42.2\% in micro average\footnote{Micro average aggregates the outcomes of all classes to compute the metrics, heavily weighing the classes with more instances. This method is effective for evaluating the overall effectiveness of the model, particularly in datasets where some classes significantly outnumber others.} and 55.4\% in macro average\footnote{Macro average calculates individual metrics for each class without regard to class frequency, and then averages these metrics, treating all classes with equal importance. This approach highlights model performance on minority classes and is beneficial for assessing model fairness across diverse class distributions.} 
A comparison with human evaluation results reveals that ORLMs outperform senior undergraduate students on simpler problems (NL4OPT/MAMO-EasyLP) and approach expert-level performance. However, for more complex problems (IndustryOR/MAMO-ComplexLP), ORLMs still underperform relative to the average student, indicating that their problem-solving capabilities for complex tasks remain inadequate. We will analyze the underlying reasons for this in Section \ref{sec: Analysis and Discussion}, highlighting that ORLMs still have the potential to surpass student performance and approach expert-level solutions on more challenging problems.

While ORLMs demonstrate strong performance in optimization modeling tasks, we also evaluate their efficacy on general large language model benchmarks after domain-specific fine-tuning. In particular, we conduct a rigorous evaluation of the LLaMA-3-8B model (pre-fine-tuning) and the ORLM-{\scriptsize LLaMA-3-8B} model (post-fine-tuning) following the open-instruct \citep{tulu} on benchmarks from various domains, including mathematics (GSM8K), code (HumanEval), and general knowledge (MMLU, BBH, TydiQA), the results are listed in Table \ref{table: ORLM in general benchmark}.

\begin{table}[!ht]
    \centering
    
    \caption{Comparison of accuracy rate between LLaMA-3-8B and ORLM-LLaMA-3-8B across multiple benchmark domains.}
    \label{table: ORLM in general benchmark}
    \begin{tabular}{l c c c c c c}
        \toprule
        \textbf{Model} & \textbf{GSM8K} & \textbf{HumanEval} & \textbf{MMLU} & \textbf{BBH} & \textbf{TydiQA} \\
        \midrule
        LLaMA-3-8B & 56.5\%  & 67.7\% & 65.2\% & 63.7\% & 21.1\% \\
        ORLM-{\scriptsize LLaMA-3-8B} & 58.0\%  & 70.6\% & 64.6\% & 61.9\% & 21.9\% \\
        \bottomrule
    \end{tabular}
\end{table}

The results reveal that ORLM-{\scriptsize LLaMA-3-8B} exhibits improved performance in the mathematics and coding domains, which aligns with expectations, as the OR-Instruct data emphasizes mathematical modeling and coding with solver assistance. However, the model exhibits a slight decline in general domain performance. Overall, the average performance across all domains shows no substantial deviation.

{

\textbf{A comparative study on problem-solving efficiency.} To verify the effectiveness of ORLM in human–AI collaboration, we conduct a comparative experiment to demonstrate its impact on algorithm engineers in solving practical problems. Specifically, we recruit a group of 30 people, consisting of experts and senior undergraduate students, dividing them evenly into two groups, A and B. Each group comprises 7 experts and 8 undergraduate students, with comparable proficiency levels. We design 7 problems with varying difficulty levels, ranging from textbook exercises to those approximating industrial complexity. Group A is tasked with independently modeling and programming the solutions without the aid of ORLM, while Group B utilizes ORLM for problem-solving and programming. We assess the effectiveness of ORLM based on two dimensions: accuracy and solution time. The results are presented in Figure \ref{fig: comparative study}. We acknowledge that the sample size of the experiment is limited due to the difficulty of finding suitable subjects; nevertheless, we perform statistical analysis on the limited data and find that the comparison is still statistically significant. In the following, we detail our statistical analysis.

\begin{figure}[!ht]    
\centering            
\subfloat[Comparison of solution time\label{fig:accuracy_time}]{
	\includegraphics[width=0.48\textwidth]{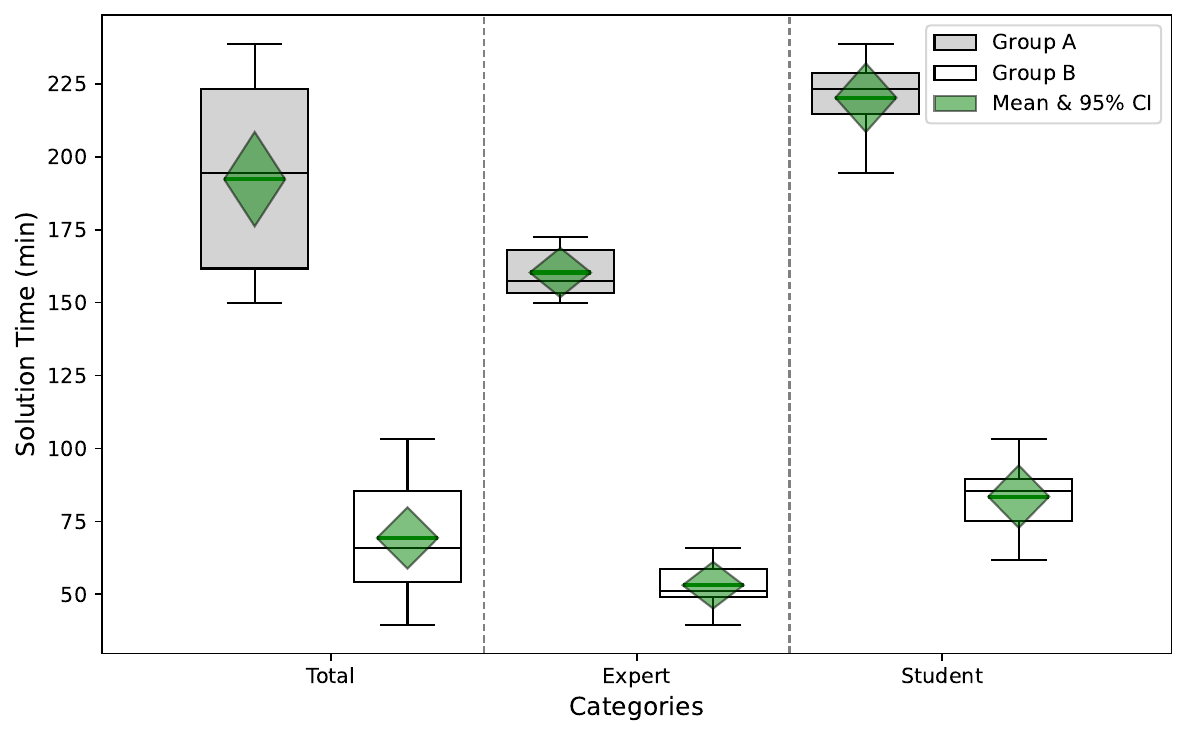}
}
\hspace{-2mm}
\subfloat[Comparison of accuracy\label{fig:orlm_usage}]{
\includegraphics[width=0.48\textwidth]{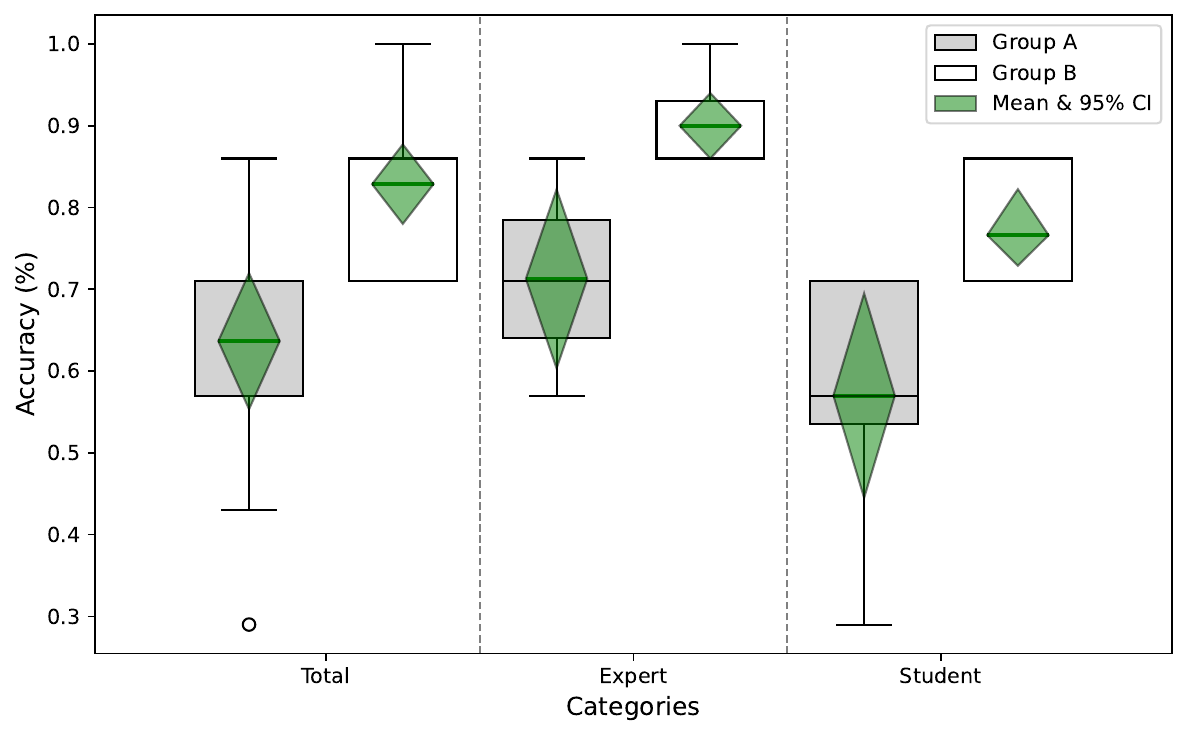}
}
\caption{A comparative study on problem-solving efficiency}
\label{fig: comparative study}
\end{figure}

%Figure \ref{fig: comparative study} presents box plots illustrating the experimental results, including overall performance as well as the two subgroups: experts and students. The green diamond markers represent the sample means along with their corresponding 95\% confidence intervals. From the figure, it is evident that Group B, assisted by ORLM, achieves approximately a 20\% increase in accuracy while requiring only one-third of the time taken by Group A to solve the same set of problems. 

Table \ref{tab:stat_results} shows the results of our statistical analysis. Specifically, for each metric (difference in accuracy for total/experts/students, difference in time for total/experts/students), we first select an appropriate statistical test based on the distributional properties of the data, as indicated in the third column of the table. For solution time of each subgroup, we find that the distribution successfully passes the normality test and meets the assumption of homogeneity of variance. Consequently, we employ a standard t-test for statistical analysis. In contrast, for accuracy, the distribution does not conform to normality; therefore, we utilize the Mann-Whitney U test as a non-parametric alternative (see \citealt{hogg2013introduction} for reference). Subsequently, we compute the test statistic and the corresponding \( p \)-value for each test, which are reported in the 4th and 5th columns, respectively. We also calculate the 95\% confidence interval for the difference between the two groups based on each test and present the results in the 6th column.

To further demonstrate the robustness of our results, the last two columns provide additional analyses, including the effect size (Hedge’s g) and the statistical power \citep{cohen2023statistical}. Notably, Hedge’s g is independent of sample size, making it a more objective measure of the generalizability and strength of the statistical significance. Meanwhile, statistical power serves as an indicator of the reliability and accuracy of our results. The detailed testing procedures and statistical indicators are provided in Appendix \ref{appendix: hypo testing}.

\begin{table}[!htbp]

  \centering
  \caption{Summary of  tests and effect sizes}
  \label{tab:stat_results}
    \resizebox{\textwidth}{!}{
  \begin{threeparttable}
  \begin{tabular}{c c c c c c c c }
    \toprule
    \multirow{2}{*}{Type} & \multirow{2}{*}{Test} & \multirow{2}{*}{Method} & \multicolumn{5}{c}{Statistic Testing} \\
    \cmidrule(lr){4-8}
          & & &Stat & $p$-value & Confidence Interval & Hedge's g & Stat Power \\
           \midrule
    \multirow{2}{*}{Total} & Accuracy & Mann-Whitney U test & 192.00 & 0.001 & (0.14, 0.24) & 1.46 & 0.97 \\
    \cmidrule(lr){2-8}
          & Time& Mann-Whitney U test & 0.00 & 0.000 & (-134.02, -112.35) & -4.43 & 1.00 \\
    \midrule
    \multirow{2}{*}{Expert} & Accuracy & Mann-Whitney U test & 44.00 & 0.008 & (0.10, 0.25) & 1.81 & 0.87 \\
    \cmidrule(lr){2-8}
          & Time& t-test & -22.50 & 0.000 & (-117.61, -96.85) & -11.26 & 1.00 \\
    \midrule
    \multirow{2}{*}{Student} & Accuracy& Mann-Whitney U test & 56.5 & 0.006 & (0.14, 0.26) & 1.56 & 0.82 \\
    \cmidrule(lr){2-8}
          & Time & t-test & -20.27 & 0.000 & (-151.29, -122.34) & -9.58 & 1.00 \\
    \bottomrule
  \end{tabular}
     \begin{tablenotes}
    \footnotesize
        \item[a)] For the criteria used in selecting between the \textit{Mann-Whitney U test} and the \textit{independent samples t-test}, please refer to Appendix \ref{appendix: hypo testing}.
    \end{tablenotes}
  \end{threeparttable}
  }
\end{table}

For Table \ref{tab:stat_results}, we can see that at a 95\% confidence level, the use of ORLM significantly improves both solution time and accuracy across all populations (total, expert, and student samples). The confidence intervals suggest that, after implementing ORLM, accuracy improves by approximately 10\%–25\%. Additionally, the total time saved in solving optimization modeling problems is 1.8–2.2 hours, with experts saving approximately 1.5–2 hours and students saving around 2–2.5 hours. The absolute value of Hedge’s g $> 1$ indicates a substantial effect size, highlighting a large difference between experimental and control groups. Additionally, statistical power $> 0.8$, ensuring a high probability of detecting a true effect and minimizing Type II error risk.

In addition, we collect feedback from engineers on their experience using ORLM, as shown in Table \ref{tab:effectiveness}. None of the engineers rate ORLM as "useless" or "minimal help". Instead, the majority provide highly positive evaluations ("significant help" or "highly effective"), indicating that ORLM is both efficient and effective in assisting their work. These demonstrate the significant potential of ORLM to enhance the application of operations research in industry through human–AI collaboration. We will further discuss the application value of ORLM in Section \ref{sec: Potential Applications}.

\begin{table}[!htbp]
  \centering
  
  \caption{Feedback from participants on ORLM effectiveness}
  \label{tab:effectiveness}
    \begin{tabular}{l c c c c c}
      \toprule
      Category & Useless & Minimal Help & Moderate Help & Significant Help & Highly Effective \\
      \midrule
      Percentage (\%) & 0.0 & 0.0 & 13.3 & 60.0 & 26.7 \\
      \bottomrule
    \end{tabular}
\end{table}

}

\subsection{Customization Enhancement}

In this subsection, we illustrate the capability of our framework to enhance large models in specific areas by generating targeted types of optimization modeling data. Specifically, we focus on the scenario of mixed integer linear programming (MILP). We provide an empirical example demonstrating how the quality of responses improves with training. Initially, we use a base dataset comprising 10,000 general modeling cases. To enhance the ORLM's proficiency in MILP, we selected 50 fundamental MILP questions and their answers for inclusion in the seed data pool. After several iterative rounds, this process yields 2,000 tailored training data points for MILP. We will subsequently assess the impact of incorporating these 2,000 data points into the training dataset by comparing the ORLM's performance before and after their addition. 
\begin{figure}[!ht]
    \centering
    \includegraphics[width=1\linewidth]{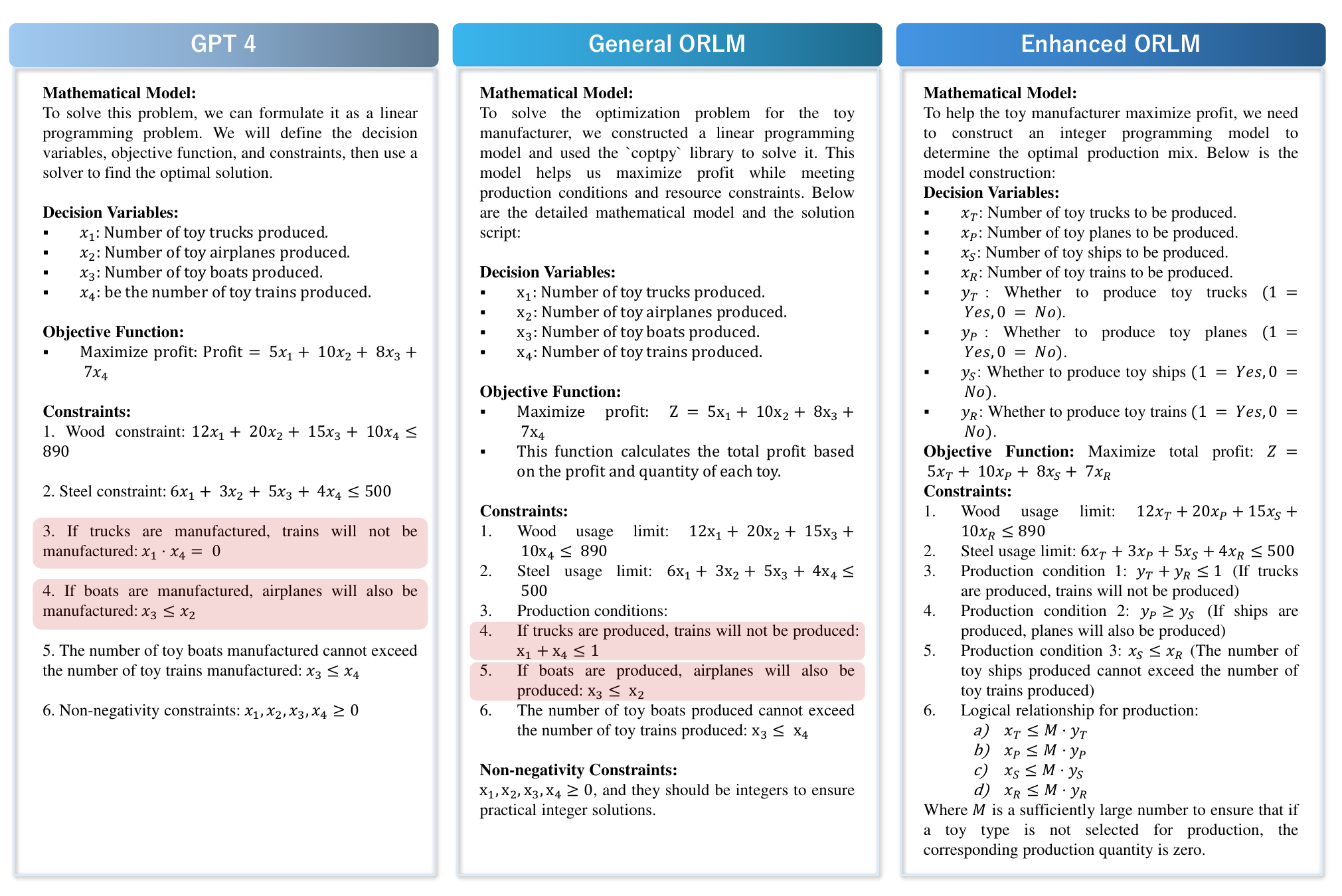}
    \caption{Comparison of ORLM before and after adding specific data for training and GPT-4 responses, the code part is omitted here. Incorrect responses are marked in red.}
    \label{fig:MILP comparsion}
\end{figure}

As depicted in Figure \ref{fig:MILP comparsion}, in response to the scenario presented in Figure \ref{fig:chat}, GPT-4 inaccurately interprets logical relationships as quantitative ones and includes nonlinear terms during the modeling process as shown in the red area, adversely affecting solver performance. A similar mistake is observed in the general ORLM, which also misinterprets logical relationships as quantitative. In contrast, the targeted-trained ORLM accurately transforms the problem into a correct mathematical model by utilizing indicator variables and the Big $M$ method. This showcases the robust generalization capability of {\sc OR-Instruct}. Furthermore, {\sc OR-Instruct} not only facilitates customized improvements across different modeling types but also demonstrates potential for targeted enhancements within specific industrial sectors.

\section{Analysis and Discussion}
\label{sec: Analysis and Discussion}

\subsection{Detailed Comparison of ORLM vs GPT-4 on IndustryOR}
\begin{figure}[!htbp]
    \centering
    \includegraphics[width=1\linewidth]{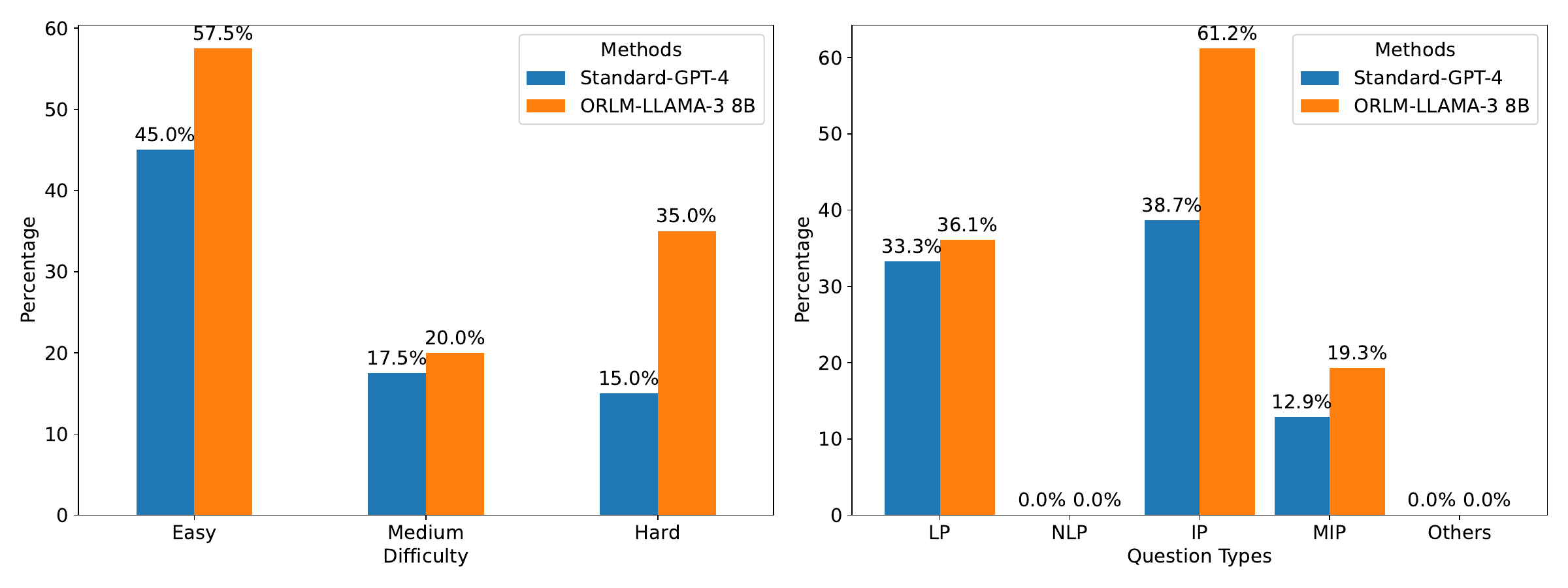}
    \caption{Accuracy rate of ORLM and GPT-4 on IndustryOR across different difficulty levels and question types.}
    \label{fig:compare ORLM and GPT-4 on IndustryOR}
\end{figure}

To assess the optimization modeling capabilities across different levels of difficulty and question types, we compare the accuracy rate of ORLM-{\scriptsize LLaMA-3-8B} and Standard-{\scriptsize GPT-4} on IndustryOR, as shown in Figure \ref{fig:compare ORLM and GPT-4 on IndustryOR}. ORLM-{\scriptsize LLaMA-3-8B} shows superior performance over Standard-{\scriptsize GPT-4} across all difficulty levels, especially in the hard category. Regarding different question types, ORLM-{\scriptsize LLaMA-3-8B} outperforms Standard-{\scriptsize GPT-4} in linear programming, integer programming, and mixed-integer programming. Both models perform poorly in non-linear programming and other rare question types. This is partly due to the fact that the seed dataset collected contains too few examples of nonlinear and other types of optimization modeling, resulting in a low percentage of that type in the training dataset, and partly due to the fact that the inherent complexity and scarcity of these types of questions. Overall, the {\sc OR-Instruct} data proves effective in enhancing \textit{Comprehensive Coverage} across various question types and difficulty levels.

\subsection{Ablation Study}

\subsubsection*{Ablation Study on Data Synthesis Strategy}

{

    We first assess whether the data generated by OR-Instruct truly leads to improvements in the Seed data, particularly in terms of the performance of the ORLM. Particularly, we compare the performance of fine-tuning on seed data alone with the results of fine-tuning on OR-Instruct-generated data based on the same seed data. We observe that, with the exception of minor variations in simpler linear programming cases (MAMO\_EasyLP), the OR-Instruct-generated data yield a significant performance boost for ORLM than only using the seed data as shown in Table \ref{table2:data_comparison}. 

\begin{table}[!htbp]
\centering

\caption{Performance comparison with varying data sizes}
\label{table2:data_comparison}
\small{\begin{tabular}{lc|cccc|ccc}
\toprule
\textbf{Base Model} & \textbf{Data Size} & \textbf{NL4OPT} & \makecell[c]{\textbf{MAMO}\\\textbf{EasyLP}} & \makecell[c]{\textbf{MAMO}\\\textbf{ComplexLP}} & \textbf{IndustryOR} & \makecell[c]{\textbf{Micro}\\\textbf{Avg}} & \makecell[c]{\textbf{Macro}\\\textbf{Avg}}\\
\midrule
\makecell[l]{LLaMA-3-8B} & {\footnotesize 32,481 (Full data)} & 85.7\% & 82.3\% & 37.4\% & 38.0\% & 71.4\% & 60.8\% \\ 
\makecell[l]{LLaMA-3-8B} & {\footnotesize 686 (seed data)} & 75.1\% & 82.8\% & 29.4\% & 20.0\% & 67.0\% & 51.8\% \\ 
\bottomrule
\end{tabular}}
\end{table}

}

\subsubsection*{Ablation Study on {\sc OR-Instruct} Augmentations}

To further verify the effectiveness of the augmentations in {\sc OR-Instruct}, we conduct detailed ablation experiments to study the effects of altering objectives and constraints, rephrasing questions, and incorporating multiple modeling techniques. {Specifically, we construct four distinct datasets, each comprising 3,000 instances, drawn from the {\sc OR-Instruct} data but employing different augmentation strategies. These datasets are then utilized to train the LLaMA-3-8B model, maintaining consistent hyperparameters across all experimental conditions.} The results are presented in Table \ref{tab:ablation}. Training data with all three augmentations (denoted as Full Augmentations) achieves a base performance of 68.6\% in micro average and 55.7\% in macro average. Removing any of the three augmentations from the base setting leads to a performance drop across all benchmarks, both in micro and macro averages. Rephrasing questions seems slightly more important than the other two. Overall, the results show that all three augmentations contribute to general performance, proving their effectiveness in enhancing the \textit{data diversity}.

\begin{table}[!htbp]
\centering
\caption{Ablation study on {\sc OR-Instruct} augmentations.}
\label{tab:ablation}
\small{\begin{tabular}{lcccccc}
\toprule
\textbf{Method} & \textbf{NL4OPT} & \makecell[c]{\textbf{MAMO}\\\textbf{EasyLP}} & \makecell[c]{\textbf{MAMO}\\\textbf{ComplexLP}} & \textbf{IndustryOR} & \makecell[c]{\textbf{Micro}\\\textbf{Avg}} & \makecell[c]{\textbf{Macro}\\\textbf{Avg}}\\
\midrule
\rowcolor{gray!17} \makecell[l]{Full Augmentations}   & 78.3\% & 80.6\% & 43.1\% & 21.0\% & 68.6\% & 55.7\% \\ 
\midrule
\makecell[l]{w/o Altering Obj\&Const} & 77.5\% & 79.2\% & 36.4\% & 20.0\% & 66.4\% & 53.2\% \\
\makecell[l]{w/o Rephrasing Questions}  & 74.2\% & 77.3\% & 41.1\% & 15.0\% & 65.1\% & 51.9\% \\
\makecell[l]{w/o Multiple Modeling}   & 78.3\% & 78.0\% & 38.8\% & 18.0\% & 66.2\% & 53.2\% \\
\bottomrule
\end{tabular}}
\end{table}

{

\subsubsection*{Ablation Study on Question Types}

ORLM can be tailored to specific domains within the field of optimization, depending on its initial training dataset. \textit{A key consideration in this process is whether incorporating data from other domains is necessary.} This investigation aims to provide valuable guidance for implementing ORLM in specialized areas.

We use linear programming as a case study, comparing ORLM's performance when trained solely on linear programming data versus a mixed dataset including other problem types.
Specifically, we construct two training sets, each comprising 7,049 instances: one set exclusively consisting of pure linear programming (LP) problems, and the other comprising a diverse mixture of problem types\footnote{This dataset is comprised of  35\% linear programming, 40\% mixed integer programming, 25\% other types.}  We fine-tune the Meta-Llama3 8B model on both datasets, ensuring that hyper-parameter settings remain consistent across experiments. The test performance is summarized in Table \ref{table:question type test_results}:

\begin{table}[!htbp]
\centering

\caption{Performance of ORLM-{\scriptsize LLaMA-3-8B} on different datasets}
\label{table:question type test_results}
\small{\begin{tabular}{lc|cccccc}
\toprule
\textbf{Dataset} & \textbf{Data Size} & \textbf{NL4OPT} & \makecell[c]{\textbf{MAMO}\\\textbf{EasyLP}} & \makecell[c]{\textbf{MAMO}\\\textbf{ComplexLP}} & \textbf{IndustryOR} & \makecell[c]{\textbf{Micro} \\ \textbf{Avg}} & \makecell[c]{\textbf{Macro} \\ \textbf{Avg}} \\
\midrule
\makecell[l]{LP Only} & {\footnotesize 7,049} & 82.0\% & 77.4\% & 26.1\% & 20.0\% & 65.4\% & 51.4\% \\
\makecell[l]{Mixed} & {\footnotesize 7,049} & 86.5\% & 81.2\% & 34.6\% & 21.0\% & 69.8\% & 55.8\% \\
\bottomrule
\end{tabular}}
\end{table}

As illustrated in Table \ref{table:question type test_results}, even with identical test sizes, the inclusion of mixed problem types significantly enhances the performance of the ORLM-{\scriptsize LLaMA-3-8B} across various datasets. Notably, when examining the tests focused on linear programming, specifically MAMO-EasyLP and MAMO-ComplexLP, it is evident that integrating additional problem types is more effective in enhancing the ORLM’s performance on challenging linear programming problems. A plausible explanation for this phenomenon is that the incorporation of a broader variety of problems effectively strengthens the generalization ability of the ORLM-{\scriptsize LLaMA-3-8B}, allowing it to capture a wider modeling capacity and consequently improve modeling performance (See \citealt{dong2023abilities}).

\subsection{Scaling Laws}

The concept of scaling laws is a key element in the development of large models. It refers to the phenomenon that the performance of large language models improves as their size and complexity increase. In this section, we will examine the impact of scaling laws on ORLMs from two dimensions: training data and model size.

From the perspective of training data, we begin with seed data and subsequently randomly sample a predetermined number of training data from the data generated by OR-Instruct. Using the LLaMA-3-8B model as the base, we conduct training while keeping all other hyperparameters constant (as in Table \ref{tab:ablation0}), in order to observe the performance of ORLM-LLaMA-3-8B on various test sets. Regarding model size, we focus on the Qwen large model series, as it offers a more refined range of model sizes that are compatible with our hardware limitations (0.5B, 1.5B, 3B, 7B, 14B). We fix the training dataset to be all data generated by OR-Instruct. Under the condition of identical training hyperparameters, we observe the performance of different Qwen2.5 model sizes on various test sets. In our study, we use the overall macro average and micro average across four test sets as metrics to evaluate the performance of the ORLMs, as visualized in Figure \ref{fig: scaling law}. More detailed results are provided in Appendix \ref{app: scaling law experiment results}.

\begin{figure}[!ht]    
\centering            
\subfloat[Different sizes of training sets]{
	\includegraphics[width=0.45\textwidth]{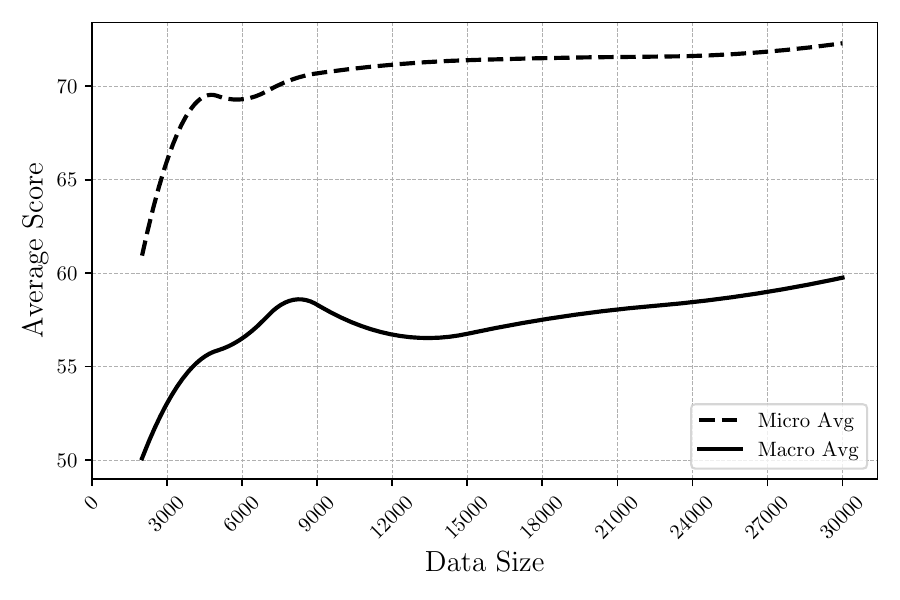}
}
\hspace{-2mm}
\subfloat[Different sizes of base models]{
\includegraphics[width=0.45\textwidth]{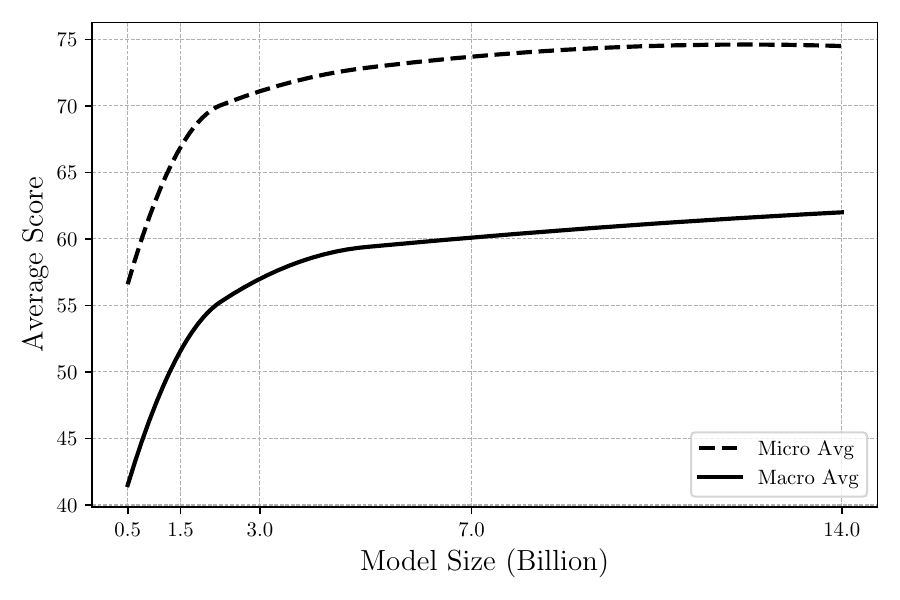}
}
\caption{Scaling law validation across different data and model sizes}
\label{fig: scaling law}
\end{figure}

The figure clearly shows an upward trend in ORLM’s accuracy on the test sets as either the model size or the data volume increases. However, the impact of these two types of scaling laws on ORLM’s performance differs. Specifically, increasing the training set size results in fluctuating performance gains from the macro average metric, this is because for simple test datasets (NL4OPT and MAMO-easyLP), ORLM quickly reaches a performance bottleneck and begins to fluctuate, causing instability in its accuracy across the entire dataset. For micro average metric, the overall performance continues to exhibit a monotonic increase with the expansion of training data, approximately following a power-law trend. This indicates that increasing the amount of training data remains beneficial for improving the model's capabilities.

Regarding the scaling law for model size, it is evident that increasing the model size significantly enhances ORLM’s accuracy. This increase not only yields steady incremental gains but also follows a power-law trend, consistent with the findings reported by \cite{Kaplan2020Scaling}. This indicates that the intrinsic capability of the base model establishes a higher performance ceiling for ORLMs in optimization modeling tasks.

\subsection{Inference Techniques}
\label{sec: Inference Techniques}

In previous model evaluations, we employed a greedy inference strategy to ensure the replicability and fairness of the results. However, using greedy sampling entails always selecting the word with the highest probability from the large model's output during each generation. While this approach guarantees deterministic results, it also limits the model's diversity and creativity. To address this, in this section, we will utilize more common sampling techniques to assess the performance improvements of the ORLMs. Specifically, we adjust the following two parameters: First, the \textit{temperature} parameter controls the diversity of the generated text. Higher temperature values increase the randomness and creativity of the generated content, while lower values tend to produce more deterministic text. Second, the \textit{top-P} parameter restricts the range of vocabulary choices considered by the model during generation. A higher \textit{top-P} value means the model will consider a broader set of candidate words, thus enhancing the diversity of the generated content. To avoid excessive randomness that may lead to significant fluctuations in the accuracy of the generated results, we apply \textit{Pass@$k$} ($k=2,4,8$) strategy to enhance stability. Specifically, for each query, the large model generates top $k$ candidate answers, and as long as at least one of the top $k$ answers is correct, it is considered to have answered this question correctly (Pass@$k$). This approach helps improve the consistency and stability of the generated outcomes. The test results is detailed in Table \ref{table:decoding_methods}.

\begin{table}[!htbp]
\centering

\caption{Performance of ORLM-LLaMA-3-8B on different decoding methods}
\label{table:decoding_methods}
\small{\begin{tabular}{lc|cccc|cc}
\toprule
\textbf{Model} & \makecell[c]{\textbf{Inference} \\ \textbf{Settings}} & \textbf{NL4OPT} & \makecell[c]{\textbf{MAMO}\\\textbf{EasyLP}} & \makecell[c]{\textbf{MAMO}\\\textbf{ComplexLP}} & \textbf{IndustryOR} & \makecell[c]{\textbf{Micro} \\ \textbf{Avg}} & \makecell[c]{\textbf{Macro} \\ \textbf{Avg}} \\
\midrule
\makecell[l]{\scriptsize ORLM-LLaMA-3-8B} & \makecell[c]{\small pass@1, \\ temp=0, \\ top-P=1.0} & 85.7\% & 82.3\% & 37.4\% & 38.0\% & 71.4\% & 60.8\% \\
\makecell[l]{\scriptsize ORLM-LLaMA-3-8B} & \makecell[c]{\small pass@2, \\ temp=0.7, \\ top-P=0.95} & 88.6\% & 83.7\% &  49.8\% &  40.0\% & 75.6\% & 65.5\% \\
\makecell[l]{\scriptsize ORLM-LLaMA-3-8B} & \makecell[c]{\small pass@4, \\ temp=0.7, \\ top-P=0.95} & 91.4\% & 85.9\% &  56.9\% &  44.0\% & 78.9\% & 69.6\% \\
\makecell[l]{\scriptsize ORLM-LLaMA-3-8B} & \makecell[c]{\small pass@8, \\ temp=0.7, \\ top-P=0.95} & 93.0\% & 88.4\% &  72.1\% &  49.0\% & 83.6\% & 75.6\% \\
\bottomrule
\end{tabular}}
\end{table}

Significant improvements have been observed in ORLMs across four test sets, with the overall micro average of Pass@8 increasing by 17.09\% relative to the previous version, while the macro average rose by 24.34\%. A comparison with Table \ref{table:main} reveals that, in terms of both macro average and micro average accuracy, the ORLM under the Pass@8 strategy has already surpassed the accuracy of senior undergraduate students and is approaching that of experts. This highlights the significant potential of ORLMs to serve as a viable alternative to human engineers. The high Pass@8 score indicates that the correct answer is often found within the top 8 solutions. However, the model's greedy decoding strategy remains insufficiently effective to consistently prioritize the optimal answer. The notable progress between the Pass@1 and Pass@8 scores suggests that while the model has the potential to generate correct solutions, its ranking capability is currently inadequate to reliably place the optimal solution in the first position. Fortunately, reinforcement learning may offer a promising solution to bridge the gap between Pass@1 and Pass@8, we will discuss it in the future work (See Section \ref{sec: Future Research Directions}).

\subsection{Limitations Analysis}

Building upon the previous discussion and results, this section concludes with an analysis of the current limitations of ORLM. First, using ORLM-{\scriptsize Llama-3-8B} as a case study, we present a statistical breakdown of error types under the greedy inference strategy, as shown in Table \ref{table:error_frequency}.

\begin{table}[!htbp]
\centering

\caption{Error frequency across different datasets}
\label{table:error_frequency}
\small{\begin{tabular}{c|cccc|ccc}
\toprule
\textbf{Error Type} & \textbf{NL4OPT} & \makecell[c]{\textbf{MAMO} \\ \textbf{Easy}} & \makecell[c]{\textbf{MAMO} \\ \textbf{Complex}} & \textbf{IndustryOR} & \makecell[c]{\textbf{Overall} \\ \textbf{Percentage}} & \makecell[c]{\textbf{Error} \\ \textbf{Percentage}} \\
\midrule
\makecell[l]{Code Error} & 1.04\% & 0.00\% & 16.59\% & 31.00\% & 5.51\% & 19.77\% \\
\makecell[l]{Model Error} & 12.46\% & 17.64\% & 45.97\% & 31.00\% & 22.28\% & 80.23\% \\
\bottomrule
\end{tabular}}
\end{table}

In the table above, the "Overall Percentage" represents the occurrence frequency of each error type across all test sets, while the "Error Percentage" indicates the proportion of each error type within the total number of errors. The data reveals that, ORLM has demonstrated considerable proficiency in utilizing COPT, achieving a pass rate of approximately 95\% across more than 1,000 test cases. Consequently, the primary challenges lie in the optimization modeling phase of the process.

To analyze the types of errors occurring in ORLM modeling, we randomly selected 100 problems from those where ORLM failed. Following manual evaluation and statistical analysis, and after excluding errors caused by coding issues, we categorized the modeling-related errors into three primary types, as detailed in Table \ref{tab:main_mistakes}. Notably, the predominant cause of ORLM's modeling errors is "Low Model Completeness", which accounts for more than half of the observed errors. Further analysis reveals that this issue stems from the limitations in the expressive and learning capabilities of 8B-scale models. These models often produce overly simplified outputs in complex scenarios, reducing the intricacy of optimization modeling problems even when such scenarios are included in the training set. This highlights the importance of fine-tuning larger-scale models, a conclusion supported by our scaling law experiments.

\begin{table}[!htbp]
\centering

\caption{Analysis of main mistakes in optimization modeling}
\label{tab:main_mistakes}
\small{\begin{tabular}{cp{6cm}c} % 限制Description列宽度为8cm
\toprule
\textbf{Modeling Error Type} & \qquad\qquad\textbf{Description} & \textbf{Proportion}  \\
\midrule
\makecell[c]{Semantic Misunderstanding} & Misunderstand the problem that the optimization model needs to solve. & 13.40\% \\
\makecell[c]{Errors in Objective/\\Constraint Translation} & Correctly understand the requirements, but make errors when formulating the optimization model. & 30.30\% \\
\makecell[c]{Low Model Completeness} & Ignore some constraints or implicit conditions of the real problem, simplifying the complex problem. & 56.30\% \\
\bottomrule
\end{tabular}}
\end{table}

Although ORLM still has certain limitations, as demonstrated by previous experimental results, we recognize substantial opportunities for improvement through the application of scaling laws, different inference techniques and others. ORLM is still promising in addressing complex optimization modeling problems, even certain real-world industrial problems within specific domains. In the next section, we will demonstrate how ORLM can be applied to practical scenarios such as industrial production, education, and other real-world applications.

}

{

\section{Future Direction}

In previous sections, we conduct a detailed analysis of ORLM's strengths and limitations within the field of optimization modeling. In this section, we build on above experimental results to discuss the practical applications of ORLM and potential directions for future improvement.

\subsection{Potential Applications}
\label{sec: Potential Applications}

Large language models are increasingly influential across various industries, and the rise of privatized and domain-specific LLMs presents new topics for management. This work represents the first attempt to fine-tune open-source LLMs specifically for optimization modeling. The ORLM not only surpasses the performance of cutting-edge proprietary models but also ensures privacy, security, and stability for commercial applications. Additionally, its 7B size allows for private deployment on personal computers with only 16GB of VRAM, making operations research more accessible to a wider audience. We hope that this work could shed lights on the deep integration of large language models with operations research and management science. Potential applications are listed as follows:

\begin{enumerate}
    \item \textbf{Revolutionizing the role of operations research in industry}: In recent years, operations research has rapidly permeated various industrial sectors, supporting companies in making more informed and effective decisions. However, optimization modeling tasks in industry often encounter two critical challenges: (1) optimization models are typically not robust enough to withstand changes in external conditions. When the environment shifts, models must be quickly adjusted to help companies make timely decisions under new circumstances; (2) optimization projects typically rely on business experts to specify goals and constraints, while algorithm engineers are responsible for developing and solving the models. For many specialized industrial problems, communication between these two roles can be costly. Algorithm engineers may struggle to grasp the intricacies of business logic, and business experts often lack technical modeling expertise.

    The introduction of ORLM could help address many of these challenges. Particularly, companies with extensive training data can customize ORLMs to their domain based on OR-Instruct. As demonstrated in Tables \ref{table:main} and \ref{table: ORLM in general benchmark}, ORLM significantly enhances optimization modeling capabilities while retaining the broad knowledge acquired during pre-training. This allows it to quickly grasp industry-specific terminology and facilitate effective communication between business experts and algorithm engineers. ORLM operates with remarkable efficiency, generating an initial solution within seconds that both engineers and business teams can refer to. Given ORLM’s strong performance on Pass@8 metrics, it can also produce multiple solutions, supporting selection and comparison, which accelerates project iterations. Two representative workflows are illustrated in Figure \ref{fig: ORLM in enterprise}.

    \begin{figure}[!htbp]    
    \centering            
    \subfloat[\small In the course of enterprise development, the constraints and objectives of operational contexts continually evolve, necessitating swift adjustments to existing algorithmic models to meet business demands.]{
        \includegraphics[width=0.8\textwidth]{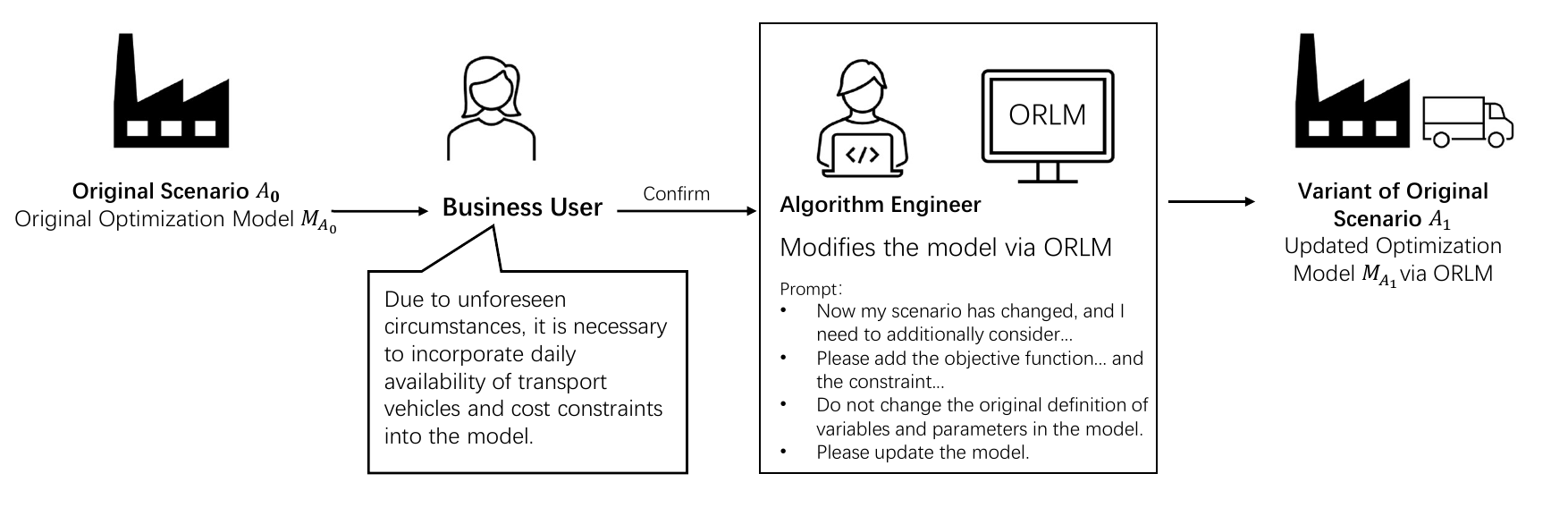}
    }
    \hspace{-2mm}
    \subfloat[\small For newly arising decision-making problems, enterprises can leverage ORLM for rapid modeling and experiment with small data samples. This approach enables enterprises to respond swiftly to market changes.]{
    \includegraphics[width=0.8\textwidth]{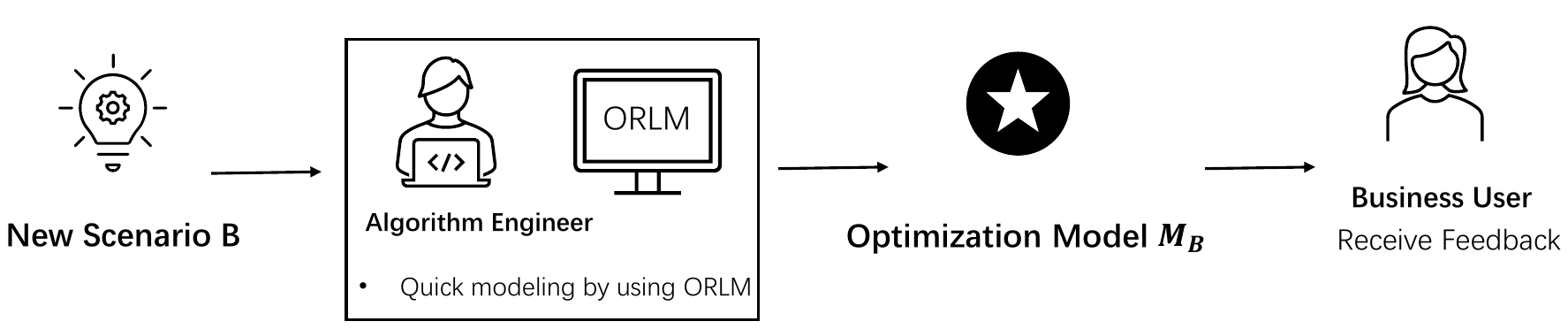}
    }
    \caption{Utilize ORLM for adaptive modeling in enterprise decision-making.}
    \label{fig: ORLM in enterprise}
    \end{figure}

    Moreover, ORLM can seamlessly modify existing optimization models by adding new business constraints, creating preliminary models, and generating initial solution code, which engineers can then validate. This capability not only reduces delivery costs but also improves accuracy on complex tasks.

    Beyond these capabilities, ORLM has the potential to leverage Retrieval-Augmented Generation (RAG) and similar technologies to incorporate internal company documents, evolving into a domain-specific optimization expert well-versed in the company’s field. This feature allows ORLM to support knowledge transfer and updates, transforming traditional project reports into valuable data assets for future model training. In addition to project delivery, ORLM can also be applied in training junior algorithm engineers in modeling techniques, further enhancing its value within the enterprise.
    \item \textbf{Operations research education}: The outstanding performance of ORLM in fundamental optimization modeling problems makes it sufficiently capable of addressing the modeling challenges encountered by beginners in operations research. Instructors can utilize ORLM for efficient instructional guidance, allowing students to quickly derive reference solutions and processes for unfamiliar operations research modeling problems, thereby enhancing teaching efficiency and quality.
    \item \textbf{Mathematical modeling competitions}: As one of the most influential competitions, mathematical modeling contests aim to cultivate students' abilities to solve problems using mathematical knowledge. Since 2023, several mathematical modeling competitions have begun to acknowledge the significant role of generative AI in modeling, permitting its use as an aid (e.g., MCM 2024). The capabilities of ORLM position it as a powerful tool in such competitions.
    \item \textbf{Reducing the learning cost of solvers}: Solvers are a core component of optimization modeling problems, yet the diverse programming syntax of different solvers significantly increases the learning cost for users, hindering their widespread adoption. The emergence of ORLM marks a pivotal shift in this regard. Notably, Table \ref{table:error_frequency} indicates that ORLM rarely encounters errors due to coding issues (95\% accuracy rate), even with less common syntax such as that specific to COPT. This suggests that algorithm engineers can confidently develop reliable operations research models and rely on ORLM to translate them into corresponding solver code, alleviating the need to extensively learn solver syntax. 
\end{enumerate}
}

\subsection{Future Research Directions}
\label{sec: Future Research Directions}

In Section \ref{sec: Analysis and Discussion}, we conduct extensive experiments to analyze the current limitations of ORLM and to investigate the underlying causes. Based on our findings, we propose directions for future improvements, calling for community attention and collaboration to drive progress in this field:

\begin{itemize}
{ 
    \item \textbf{Incorporating sophisticated techniques:} In Section \ref{sec: Inference Techniques}, we observe that while ORLM can generate correct answers, it struggles with ranking them effectively. Incorporating reinforcement learning techniques could help narrow the gap between pass@1 and pass@8 scores. As noted by \cite{sutton2018reinforcement}, one of the primary goals of reinforcement learning is to align the policy with the Bellman optimality policy, effectively optimizing ranking quality. In this context, relevant methods like reinforcement learning from human feedback (RLHF), direct preference optimization (DPO), and Kahneman-Tversky optimization (KTO) aim to minimize the disparity between the model's current greedy decoding policy and the Bellman optimality policy, thereby enhancing the model’s ranking accuracy and moving pass@1 closer to pass@8. In addition, approaches such as multi-agent collaboration and internal chain of thought structures similar to o1 \citep{openai2024o1} could further enhance the performance of current large models in optimization modeling tasks. 
    \item \textbf{Dataset construction}: While this paper presents a data synthesis strategy, it is insufficient for reinforcement learning techniques like RLHF, which require training data in the form of a preference list. For each question, multiple responses must be ranked according to preference. Furthermore, in the context of optimization problems, incorporating actual optimal solutions into the training set would enable a wider range of techniques. Constructing such datasets may require user's feedback or the development of new data synthesis methods to better meet these needs.
    \item \textbf{Data mining and exploration}: As shown in Figure \ref{fig: scaling law}, we have demonstrated that scaling laws significantly benefit ORLM. However, for a fixed model size, the performance gains from additional data become increasingly marginal at later stages. This raises the question of whether it is possible to identify an optimized, minimal dataset or develop efficient data synthesis strategies that can balance ORLM's performance with the associated training costs. Existing studies (e.g., \citealt{chen2023maybe}) indicate that only a small portion of a complete dataset is often sufficient to achieve comparable performance. Therefore, developing a tailored data exploration strategy for optimization problem modeling, which minimizes the reliance on large-scale data resources for enhancing large models, represents an important direction for future research.}
\end{itemize}

\section{Concluding Remarks}
\label{conclusion}

In this paper, we propose a new path for the training of open-source large language models (LLMs) specifically tailored for optimization modeling. We characterize four critical requirements essential for the training dataset of optimization modeling LLMs and develop {\sc OR-Instruct}, a semi-automated framework for generating synthetic data that meets these specific needs. Additionally, we introduce the IndustryOR benchmark, the first of its kind in the industry. We utilize the {\sc OR-Instruct} data to train open-source LLMs with approximately 7 billion parameters, which significantly enhances their optimization modeling capabilities, achieving competitive performance across all test datasets.

This paper also identifies the current limitations of ORLM, such as its overly simplistic output for complex problems, weak ability to rank optimal solutions, and poor learning capability from data. Our experiments suggest that ORLM can be further enhanced by leveraging scaling laws or incorporating methods such as reinforcement learning (RL), potentially reaching expert-level performance. We hope these findings provide valuable insights for the future advancement of ORLM's applications in industry and open promising avenues for subsequent research.

Finally, we note that during the revision process of this paper, numerous high-performing large models have emerged, such as DeepSeek-R1 and Grok-3. While these new models demonstrate impressive capabilities, further enhancing their performance in optimization modeling requires additional refinement. The framework proposed in this paper offers a valuable reference for such improvements, including synthesizing high-quality data to strengthen modeling comprehension, or considering reinforcement learning strategies to align ranking capabilities. By leveraging our framework as a promising starting point, the inherent strengths of large models can be more effectively translated into optimization modeling capabilities.

\newpage
\bibliography{custom}
\bibliographystyle{plainnat}

\newpage
\begin{APPENDICES}

\begin{center}
 \Large \textbf{ORLM: A Customizable Framework in Training Large Models for Automated Optimization Modeling}
\end{center}

\counterwithin{proposition}{section}
\counterwithin{lemma}{section}
\counterwithin{example}{section}
\counterwithin{theorem}{section}
\numberwithin{equation}{section}
\renewcommand{\theequation}{\thesection. \arabic{equation}}
\setcounter{page}{1}

\setcounter{equation}{0}

\begin{center}
	{\LARGE \vspace{.1cm}
		%A Customizable Framework for Data Generation in Training Large Models for Automated Optimization Modeling\\ \vspace{.1cm}
  \Large \textbf{Online Supplement} 
	}
	\medskip
\end{center}

\section{Appendix}\label{sec:appendix}

\subsection{Criteria for Difficulty Level of Problem}
\label{app:criteria_of_difficulty}

{ We need to first highlight that the difficulty associated with optimization modeling is inherently challenging to quantify. Unlike solving optimization problems that can be assessed based on the number of variables, constraints, or model forms, the complexity of optimization modeling arises in enabling a large model to learn the mapping from natural language to optimized models. This difficulty primarily relates to the model's reasoning abilities and its capacity for abstracting complex problem descriptions.

Nevertheless, this study adopts a heuristic approach to assess the difficulty of modeling problems. We propose four evaluation criteria and provide illustrative examples of problem difficulty (see Appendix \ref{app:compare_easy_and_hard}). Using these criteria and specific examples, GPT-4 independently evaluates the difficulty level of each problem, ultimately categorizing them into three levels: easy, medium, and hard. The four criteria considered in this study are as follows:
}

\begin{enumerate}
    \item Problem size: This includes the number of constraints, objectives, and variables described in the natural language problem. Generally, the larger the problem, the higher the complexity of the model. {(As shown in the statistical data in Figure \ref{fig2:Average Variables and Constraints by Problem Type}, we sample 100 instances for each of the three difficulty levels from the training dataset. The average numbers of variables and constraints in these samples align with the proposed evaluation criteria.)}
    \item Complex and logical relationships: The complexity of relationships between variables and the logical structure of objective functions and constraints significantly affect modeling difficulty. Problems involving intricate logical conditions, such as "if-else" statements, "or" conditions, and nonlinear functions, increase the challenge of formulating precise mathematical models.
    \item Ambiguity and complexity of natural language description: Natural language descriptions often contain ambiguous or polysemous expressions, which complicate the task of translating them into precisely defined mathematical models. Ambiguities in wording, implicit constraints, and unclear priorities require additional clarification and assumptions for accurate modeling.
    \item Requirement for interdisciplinary knowledge: Certain natural language problems may encompass specific domain knowledge (e.g., economics, engineering, biology), necessitating the application of expertise from these fields for accurate modeling. This requires not only operations research skills but also an interdisciplinary understanding, thereby increasing the complexity of the modeling process.
\end{enumerate}

These four criteria collectively provide a comprehensive framework for evaluating the difficulty of translating natural language descriptions into mathematical models in operations research.

        \begin{figure}[!htbp]
    \centering
    \includegraphics[width=0.7\linewidth]{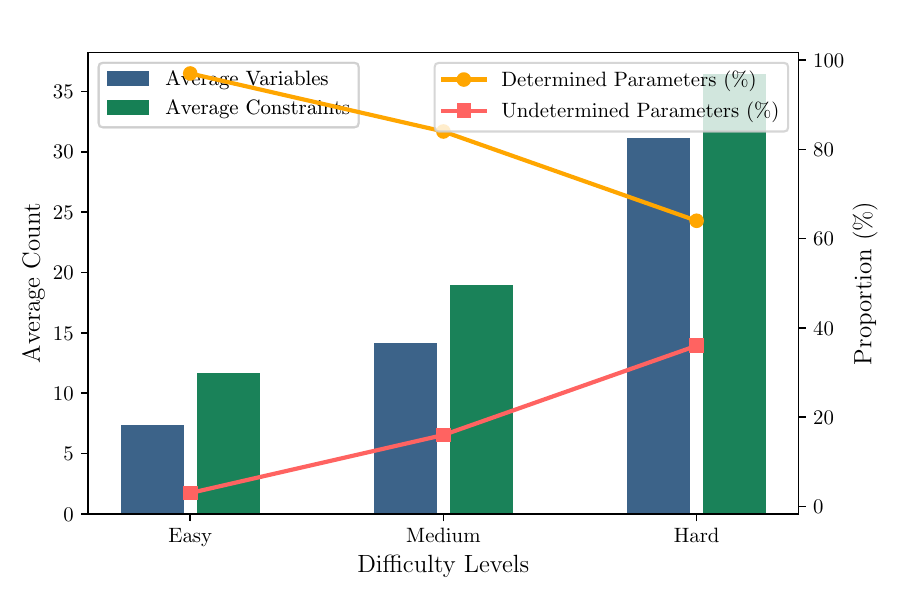}
    \caption{This figure illustrates the statistical data obtained after sampling 100 problems for each of the three difficulty levels from the training dataset. First, the proportion of optimization problems with \textbf{determined parameters} is calculated and presented by the yellow line. For this type, the average number of constraints and variables is computed and visualized as a bar chart. For problems with \textbf{undetermined parameters} (parameters such as the number of warehouses can be manually designed, refer to example provided in Appendix \ref{app: hard example}), only the proportion is calculated and represented by the red line.}
    \label{fig2:Average Variables and Constraints by Problem Type}
\end{figure}

\subsection{Prompt Template for Expansion}
\label{app:prompt_for_expansion}
{ 
\noindent\textbf{Description:  }

In the \textit{Expansion} phase, we design two prompts to enhance the effectiveness of the model. The first prompt aims to generate a sufficiently broad range of operations research scenarios and industries, serving as a reference for large language model to expand scenarios. The second prompt leverages the few-shot learning capability of the model, using three sample data points to generate a specific optimization problem within new operations research scenario. The new scenario is derived from the result produced by the first prompt. The term highlighted in the black box, \outline{select from the scenario list}, in the second prompt refers to the process of randomly selecting a scenario from the output generated by the first prompt.

}

\noindent\rule{\linewidth}{0.4pt}

\noindent \textbf{Scenario Generation Prompt:}
\texttt{\scriptsize
\\Please list 100 specific application scenarios or industries where optimization modeling are utilized. These scenarios should cover diverse domains such as logistics and supply chain management, manufacturing, energy and environment, healthcare, transportation, finance and insurance, agriculture, and military defense. Each scenario should be as specific as possible, reflecting practical uses of optimization techniques. Your results should be structured systematically and returned in JSON format.
}

\noindent\rule{\linewidth}{0.4pt}

\noindent\textbf{Prompt for Expansion:}
\texttt{\scriptsize
\\Imagine you are a seasoned operations research algorithm engineer. Your task is to develop an operations research problem that is closely aligned with real-world applications and to provide the corresponding optimization model along with COPT solver code. The problem you design should be relevant to the selected scenario and align with its specific characteristics: \outline{select from the scenario list}. Please carefully follow the format and structure of the examples provided below as a reference to craft a new optimization modeling problem.
\\
\\\# Example 1:
\\\#Scenario\#:
\\Retailing
\\
\\\#Question Type\#:
\\Integer Programming
\\
\\\#Problem\#:
\\A leather shoe store employs 5 full-time salespersons and 4 part-time salespersons. In order to optimize the working environment and consider employee health, the store decides to limit the overtime hours of each full-time employee. The following table shows the working hours and wage information for the employees:
\\
\\|  | Monthly Working Hours | Sales (pairs/hour) | Wage (dollars/hour) | Overtime Pay (dollars/hour) |
\\| :---: | :---: | :---: | :---: | :---: |
\\| Full-time | 160 | 5 | 1 | 1.5 |
\\| Part-time | 80 | 2 | 0.6 | 0.7 |
\\
\\The profit per pair of shoes sold is 0.5 dollars. The store has set the following goals:
\\
\\$p_{1}$: Achieve a monthly sales volume of at least 5500 pairs.
\\$p_{2}$: Limit the overtime hours of each full-time salesperson to no more than 20 hours.
\\$p_{3}$: Ensure full employment for all salespersons and give extra consideration to full-time employees.
\\$p_{4}$: Minimize overtime hours as much as possible.
\\
\\Please develop an objective programming model for this problem.
\\
\\\#Completion Solution\#:
\\\#\# Mathematical Model:
\\To achieve the goals of the leather shoe store, we will use an objective programming model to balance the achievement levels of each goal. The model is as follows:
\\...
\\
\\
\\\# Example 2:
\\\#Scenario\#:
\\Agriculture
\\
\\\#Question Type\#:
\\Non-Linear Programming
\\
\\\#Problem\#:
\\An agricultural company wants to optimize the climate conditions inside their greenhouse to improve the yield and quality of specific crops. The company grows two main crops: tomatoes and cucumbers. To achieve the optimal growth conditions, precise control of temperature, humidity, and CO2 concentration inside the greenhouse is required. Each crop has different requirements for these environmental factors, and the company wants to adjust the greenhouse's environmental parameters to meet the growth needs of both crops and maximize the total yield.
\\
\\\#Completion Solution\#:
\\\# Mathematical Model:
\\\#\#\# Decision Variables:
\\- \( T \): Temperature inside the greenhouse (°C)
\\- \( H \): Humidity inside the greenhouse (\%)
\\- \( C \): CO2 concentration inside the greenhouse (ppm)
\\...
\\
\\...one more example...
\\
\\\# Example:
\\
}

\subsection{Prompt Template for Augmentation}

{ 
\noindent\textbf{Description:  }

In the following steps for \textit{Augmentation}, text highlighted by black boxes, such as \outline{Input ...}, represents placeholders that need to be filled in during actual use as input for the large language model (e.g., GPT-4). Parentheses (You should provide ...) act as prompts, indicating that the model will generate improved output to be placed at the specified location.
}
\noindent\rule{\linewidth}{0.4pt}

\vspace{0.4cm}
\subsubsection{Prompt Template for Altering Objective and Constraints}
\label{app:prompt_for_altering}

\texttt{\scriptsize
\\
\\Image you are an operations research algorithm engineer, I will provide you with an entry containing a problem description, an optimization model, and COPT code. Your task is to modify the constraints or objectives in the problem description based on the actual situation and accordingly adjust the corresponding model and code. Please follow the steps below for your output:
\\
\\1. Real-World Scenario Consideration:
\\ Based on the problem description, please consider how this problem would change in a real-world scenario, listing a specific as well as feasible change.
\\
\\2. Modify Problem Description:
\\ Considering the variation you listed in the previous step, make changes to the problem description.
\\
\\3. Model and Code Modification:
\\ You are asked to modify the original operations research model based on the original operations research model, taking into account the changed problem situation, and modify the corresponding COPT code.
\\
\\4. Complete Output:
\\ After modifying the model, you should output the complete problem description, mathematical model, and COPT code for the new scenario in the original format.
\\
\\Task:
\\Follow the steps outlined to adapt the problem for a real-world scenario, and provide a complete solution in the original format.
\\
\\---
\\
\\Original Entry:
\\\[ \text{\outline{Input the original entry here}} \]
\\
\\---
\\
\\\#Completion Solution\#:
\\\#\# Modified Problem Description:
\\To address the real-world scenario identified, the problem description has been modified as follows:
\\
\\ \[\text{(You should provide the modified problem description here)}\]
\\
\\\#\# Modified Mathematical Model:
\\Based on the changes in problem scenario, the mathematical model has been adapted as follows:
\\
\\ \[\text{(You should provide the modified mathematical model here)}\]
\\
\\\#\# Modified COPT Code:
\\The COPT code has been updated to reflect the changes in the mathematical model:
\\
\\ \[\text{(You should provide the modified COPT code here)}\]
\\
}

\subsubsection{Prompt Template for Rephrasing Questions}
\label{app:prompt_for_rephrasing}

\texttt{\scriptsize
\\
\\
\\\#Task\#:
\\Image you are a user working with a large language model. I will provide you with an entry containing a problem description, an optimization model, and COPT code. Your task is to rephrase the problem description using your own wording, ensuring that your expression differs in style from the original, while preserving the original meaning. You should ensure that the mathematical model remains consistent and valid under the revised description.
\\
\\Your output should replace the original problem description with your improved version, leaving the optimization model and COPT code completely unchanged. The final output must retain the same format as the original entry. Below is the original entry:
\\
\\---
\\
\\Original Entry:
\\\[ \text{\outline{Input original entry here}} \]
\\---
\\
\\\#Completion Solution\#:
\\The rewritten problem description is as follows:
\\\[ \text{(You should provide the rewritten problem description here)} \]
\\
}

\subsubsection{Prompt Template for Incorporating Multiple Modeling Techniques}
\label{app:prompt_for_multiple}

\texttt{\scriptsize
\\
\\Task:
\\Image you are a seasonal operations research engineer, I will provide you with an entry containing a problem description, an optimization model, and COPT code. Your task is to reconstruct the optimization model, taking into account a broader range of modeling techniques.
\\
Follow the following steps to enhance the model by incorporating multiple modeling techniques. Provide detailed explanations for each modification and the rationale behind selecting each technique.
\\
\\\#Execution steps\#:
\\Incorporate multiple modeling techniques into the provided mathematical model. Follow these steps:
\\
\\1. Techniques Instruction:
\\   The following shows the different modeling techniques and the conditions under which they are applied:
\\   ...
\\   - Auxiliary Variables: Suitable for simplifying complex relationships or non-linearities in the model.
\\   - Big M Method: Appropriate for models with conditional constraints within a linear programming framework.
\\   - Penalty Functions: Useful for converting hard constraints into an unconstrained optimization problem.
\\   ...
\\
\\2. Identify Modification Needs:
\\   Analyze the original model and identify areas where modifications could be used based on Techniques Instruction.
\\
\\3. Modify the Model:
\\   Apply the selected technique(s) to modify either the objective function or the constraints of the original mathematical model.
\\
\\4. Modify the Code:
\\   Based on the modified model, the corresponding code is modified following the code format of COPT.
\\   
\\5. Organize the results:
\\   Organize the problem description, modified model and modified code strictly against the format of the original problem.
\\
\\---
\\
\\Original Entry:
\\\[ \text{\outline{Input the original entry here}} \]
\\
\\---
\\
\\\#Completion Solution\#:
\\\#\# Mathematical Model:
\\After incorporating additional modeling techniques, the reconstructed optimization model is as follows:
\\
\\   \[
\\   \text{(You should provide the modified model here)}
\\   \]
\\...
\\
\\\#\# Modified Code:
\\Based on the enhanced mathematical model, the corresponding code in COPT format is provided below:
\\
\\\[\texttt{(You should provide the modified code here)}\]
\\
\\
}

\subsection{Alpaca-like Template for ORLMs Training}
\label{app:alpaca_template}
{ 
\noindent\textbf{Description:  }

The following template is used for training and testing the ORLM. When utilizing this template, users should replace the \{Question\} section with a natural language description of the optimization modeling problem. The ORLM will then return outputs in a standardized format.
}
\noindent\rule{\linewidth}{0.4pt}

\texttt{\scriptsize
\\Below is an operations research question. Build a mathematical model and corresponding python code using `coptpy` that appropriately addresses the question.\\
\\
\# Question:\\
\{Question\}\\
\\
\# Response:\\
}

% \subsection{Hyper-parameters for Training ORLMs}
% \label{app:hyper_parameters}

\subsection{Detailed Results of Numerical Experiments}

\subsubsection{Numerical Results for Different Question Types}.

\begin{table}[!htbp]
\centering
\caption{Comparison of ORLM and GPT-4 on IndustryOR across different difficulty levels and question types.}
\label{tab:orlm_vs_gpt4}
\small{\begin{tabular}{l|ccc|ccccc}
\toprule
\textbf{Method} & \multicolumn{3}{c}{\textbf{Difficulty}} & \multicolumn{5}{c}{\textbf{Question Types}}\\
 & Easy & Medium & Hard & LP & NLP & IP & MIP & Others\\
\midrule
\makecell[l]{Standard-\scriptsize{GPT-4}} & 45.0\% & 17.5\%& 15.0\%& 33.3\% & 0.0\% & 38.7\% & 12.9\% & 0.0\% \\
\rowcolor{gray!17} \makecell[l]{ORLM-\scriptsize{LLaMA-3-8B}} & 57.5\% & 20.0\% & 35.0\% & 36.1\% & 0.0\% & 61.2\% & 19.3\% & 0.0\% \\ 
\bottomrule
\end{tabular}}
\end{table}

\subsubsection{Scaling Law Numerical Results}.
\label{app: scaling law experiment results}

~\\

\begin{table}[!htbp]
\centering
\caption{Performance comparison with varying data sizes}
\label{table:data_comparison}
\small{\begin{tabular}{lc|cccc|cc}
\toprule
\textbf{Base Model} & \textbf{Data Size} & \textbf{NL4OPT} & \makecell[c]{\textbf{MAMO}\\\textbf{EasyLP}} & \makecell[c]{\textbf{MAMO}\\\textbf{ComplexLP}} & \textbf{IndustryOR} & \makecell[c]{\textbf{Micro}\\\textbf{Avg}} & \makecell[c]{\textbf{Macro}\\\textbf{Avg}}\\
\midrule
\makecell[l]{LLaMA-3-8B} & {\footnotesize 30,000} & 88.9\% & 81.4\% & 43.6\% & 25.0\% & 72.3\% & 59.8\% \\ 
\makecell[l]{LLaMA-3-8B} & {\footnotesize 25,000} & 86.5\% & 82.4\% & 40.3\% & 25.3\% & 71.7\% & 58.6\% \\ 
\makecell[l]{LLaMA-3-8B} & {\footnotesize 20,000} & 88.2\% & 82.5\% & 37.9\% & 23.0\% & 71.6\% & 57.9\% \\ 
\makecell[l]{LLaMA-3-8B} & {\footnotesize 10,000} & 87.8\% & 81.8\% & 35.9\% & 25.0\% & 70.9\% & 57.6\% \\ 
\makecell[l]{LLaMA-3-8B} & {\footnotesize 8,000}  & 83.7\% & 80.4\% & 40.1\% & 30.0\% & 70.4\% & 58.6\% \\ 
\makecell[l]{LLaMA-3-8B} & {\footnotesize 6,000}  & 83.3\% & 79.9\% & 38.9\% & 24.0\% & 69.3\% & 56.5\% \\ 
\makecell[l]{LLaMA-3-8B} & {\footnotesize 4,000}  & 80.0\% & 82.2\% & 33.6\% & 24.0\% & 68.9\% & 55.0\% \\ 
\makecell[l]{LLaMA-3-8B} & {\footnotesize 2,000}  & 81.2\% & 68.9\% & 26.1\% & 24.2\% & 60.9\% & 50.1\% \\ 
\bottomrule
\end{tabular}}
\end{table}

\begin{table}[!htbp]
\centering
\caption{Model performance comparison with different model sizes}
\label{table: Model performance comparison with different model sizes}
\small{\begin{tabular}{lc|cccc|ccc}
\toprule
\textbf{Base Model} & \textbf{Model Size} & \textbf{NL4OPT} & \makecell[c]{\textbf{MAMO}\\\textbf{EasyLP}} & \makecell[c]{\textbf{MAMO}\\\textbf{ComplexLP}} & \textbf{IndustryOR} & \makecell[c]{\textbf{Micro}\\\textbf{Avg}} & \makecell[c]{\textbf{Macro}\\\textbf{Avg}}\\
\midrule
\makecell[c]{Qwen2.5} & \footnotesize 14B & 90.2\% & 83.5\% & 48.3\% & 26.0\% & 74.5\% & 62.0\%  \\ 
\makecell[c]{Qwen2.5} & \footnotesize 7B & 86.1\% & 85.2\% & 44.1\% & 25.0\% & 73.7\% & 60.1\%  \\ 
\makecell[c]{Qwen2.5} & \footnotesize 3B & 83.2\% & 84.2\% & 36.5\% & 24.0\% & 71.1\% & 57.0\%  \\ 
\makecell[c]{Qwen2.5} & \footnotesize 1.5B & 75.5\% & 82.5\% & 30.3\% & 18.0\% & 66.9\% & 51.6\%  \\ 
\makecell[c]{Qwen2.5} & \footnotesize 0.5B & 63.6\% & 73.3\% & 16.1\% & 13.0\% & 56.6\% & 41.5\%  \\ 
\bottomrule
\end{tabular}}
\end{table}

{
\subsection{Hypothesis Testing on the Effectiveness of ORLM}
\label{appendix: hypo testing}

As stated in the main text, a total of 30 participants were recruited for the study, including 14 experts and 16 students. The specific groupings and corresponding experimental results are presented in Table \ref{tab:Statistical table of ORLM experimental participants' data}.

\begin{table}[!htbp]

  \centering
  \caption{Statistical table of ORLM experimental participants' data}
  \label{tab:Statistical table of ORLM experimental participants' data}
  \begin{tabular}{l c D{.}{.}{2.2} D{.}{.}{2.2} c D{.}{.}{2.2} D{.}{.}{2.0}}
    \toprule
    \multirow{3}{*}{Group} & \multicolumn{3}{c}{Group A} & \multicolumn{3}{c}{Group B} \\
    \cmidrule(lr){2-4} \cmidrule(l){5-7}
          &   \multicolumn{1}{c}{Index}& \multicolumn{1}{c}{Accuracy Rate (\%)} & \multicolumn{1}{c}{Time(min)} & \multicolumn{1}{c}{Index} & \multicolumn{1}{c}{Accuracy Rate (\%)} & \multicolumn{1}{c}{Time(min)} \\
    \midrule
    \multirow{7}{*}{Expert} & AE1 & 0.86 & 149.9 & BE1 & 1.00 & 49.1 \\
          & AE2 & 0.71 & 157.4 & BE2 & 0.86 & 65.7 \\
          & AE3 & 0.86 & 166.0 & BE3 & 0.86 & 39.6 \\
          & AE4 & 0.71 & 172.4 & BE4 & 0.86 & 51.1 \\
          & AE5 & 0.57 & 151.0 & BE5 & 0.86 & 57.1 \\
          & AE6 & 0.71 & 155.4 & BE6 & 1.00 & 48.7 \\
          & AE7 & 0.57 & 170.2 & BE7 & 0.86 & 60.4 \\
    \midrule
    \multirow{8}{*}{Student} & AS1 & 0.57 & 224.3 & BS1 & 0.71 & 87.0 \\
          & AS2 & 0.71 & 230.7 & BS2 & 0.86 & 88.1 \\
          & AS3 & 0.43 & 217.3 & BS3 & 0.71 & 103.2 \\
          & AS4 & 0.71 & 222.4 & BS4 & 0.86 & 94.1 \\
          & AS5 & 0.57 & 194.5 & BS5 & 0.71 & 83.7 \\
          & AS6 & 0.29 & 238.7 & BS6 & 0.71 & 74.7 \\
          & AS7 & 0.71 & 228.0 & BS7 & 0.86 & 61.6 \\
          & AS8 & 0.57 & 206.4 & BS8 & 0.71 & 75.4 \\
    \bottomrule
  \end{tabular}
\end{table}

Our objective is to evaluate whether the use of the ORLM tool leads to significant improvements in both time efficiency and accuracy for students and experts, respectively. Accordingly, we formulate the null and alternative hypotheses as follows:
\begin{itemize}
    \item \textit{Null Hypothesis $H_0$}: There is \textbf{no significant difference} in the population means of the experimental data for experts (or students) before and after using the ORLM tool.
    \item \textit{Alternative Hypothesis $H_1$}: There is a \textbf{significant difference} in the population means of the experimental data for experts (or students) before and after using the ORLM tool.
\end{itemize}

Next, we conduct statistical tests separately for experts and students on both accuracy and time efficiency data. Before performing hypothesis testing, we first examine the distributional properties of the data to determine the most appropriate testing method. Specifically, we assess whether the data follow a normal distribution and whether homogeneity of variance is satisfied.  

To test for normality, we employ the \textit{Shapiro-Wilk test}, as it is more sensitive for small sample sizes (\( n \leq 50 \)) \citep{shapiro1965analysis}. Similarly, to assess the homogeneity of variance, we use the \textit{Levene’s test}, which is also suitable for small samples \citep{ott2010introduction}. The results of these tests are summarized in Table \ref{tab:shapiro_wilk_levene}.

\begin{table}[!htbp]
  \centering
  
  \caption{Shapiro-Wilk and Levene test results}
  \label{tab:shapiro_wilk_levene}
  \begin{tabular}{c c c c c c c}
    \toprule
    \multirow{2}{*}{Type} & \multirow{2}{*}{Test} & \multirow{2}{*}{Group} & \multicolumn{2}{c}{Shapiro-Wilk} & \multicolumn{2}{c}{Levene} \\
    \cmidrule(lr){4-5} \cmidrule(l){6-7}
          & & & stat & $p$-value & stat & $p$-value \\
          \midrule
    \multirow{4}{*}{Total} & \multirow{2}{*}{Accuracy} & Group A & 0.900 & 0.098 & \multirow{2}{*}{1.464} & \multirow{2}{*}{0.236} \\
          & & Group B & 0.801 & 0.003 & \\
    \cmidrule(lr){2-7}
          & \multirow{2}{*}{Time} & Group A & 0.879 & 0.045 & \multirow{2}{*}{11.066} & \multirow{2}{*}{0.002} \\
          & & Group B & 0.962 & 0.737 & \\
    \midrule
    \multirow{4}{*}{Expert} & \multirow{2}{*}{Accuracy} & Group A & 0.857 & 0.143 & \multirow{2}{*}{1.203} & \multirow{2}{*}{0.294} \\
          & & Group B & 0.600 & 0.000 & \\
    \cmidrule(lr){2-7}
          & \multirow{2}{*}{Time} & Group A & 0.903 & 0.350 & \multirow{2}{*}{0.103} & \multirow{2}{*}{0.754} \\
          & & Group B & 0.975 & 0.931 & \\
    \midrule
    \multirow{4}{*}{Student} & \multirow{2}{*}{Accuracy} & Group A & 0.860 & 0.120 & \multirow{2}{*}{1.201} & \multirow{2}{*}{0.292} \\
          & & Group B & 0.641 & 0.000 & \\
    \cmidrule(lr){2-7}
          & \multirow{2}{*}{Time} & Group A & 0.951 & 0.720 & \multirow{2}{*}{0.013} & \multirow{2}{*}{0.909} \\
          & & Group B & 0.983 & 0.976 & \\
    \bottomrule
  \end{tabular}
\end{table}

Table \ref{tab:shapiro_wilk_levene} presents Shapiro-Wilk and Levene test results for two outcome measures — accuracy and time — across total, expert, and student samples comparing Group A and Group B. 

While the expert and student subgroups show normality in time data for both groups (with $p$-values above $0.05$), the total sample reveals marginal normality in Group A ($p=0.045$) and significant heterogeneity of variance ($p=0.002$). For accuracy, all samples (total, expert, and student) exhibit severe non-normality in Group B ($p \leq 0.003$), despite meeting variance homogeneity criteria ($p>0.05$). Consequently:
\begin{itemize}
    \item Time: A Mann-Whitney U test is recommended for the total sample due to Group A's marginal normality and variance heterogeneity, though independent $t$ -tests may still apply to expert/student subgroups.
    \item Accuracy: A Mann-Whitney U test is used across all samples due to Group B's consistent non-normality.
\end{itemize}

We summarize the results in Table \ref{tab:statistic_testing}.

\begin{table}[!htbp]

  \centering
  \caption{Statistical testing results}
  \label{tab:statistic_testing}
  \resizebox{\textwidth}{!}{
  \begin{threeparttable}
  \begin{tabular}{c c c c c c c c }
    \toprule
    \multirow{2}{*}{Type} & \multirow{2}{*}{Test} & \multirow{2}{*}{Method} & \multicolumn{5}{c}{Statistic Testing} \\
    \cmidrule(lr){4-8}
          & & &Stat & $p$-value & Confidence Interval & Hedge's g & Stat Power \\
           \midrule
    \multirow{2}{*}{Total} & Accuracy & Mann-Whitney U test & 192.00 & 0.001 & (0.14, 0.24) & 1.46 & 0.97 \\
    \cmidrule(lr){2-8}
          & Time& Mann-Whitney U test & 0.00 & 0.000 & (-134.02, -112.35) & -4.43 & 1.00 \\
    \midrule
    \multirow{2}{*}{Expert} & Accuracy & Mann-Whitney U test & 44.00 & 0.008 & (0.10, 0.25) & 1.81 & 0.87 \\
    \cmidrule(lr){2-8}
          & Time& t-test & -22.50 & 0.000 & (-117.61, -96.85) & -11.26 & 1.00 \\
    \midrule
    \multirow{2}{*}{Student} & Accuracy& Mann-Whitney U test & 56.5 & 0.006 & (0.14, 0.26) & 1.56 & 0.82 \\
    \cmidrule(lr){2-8}
          & Time & t-test & -20.27 & 0.000 & (-151.29, -122.34) & -9.58 & 1.00 \\
    \bottomrule
  \end{tabular}
  \end{threeparttable}
  }
\end{table}

Table \ref{tab:statistic_testing} presents key statistical metrics, including the statistic value (Stat), the $p$-value (p-value), the 95\% confidence level of the difference (Confidence Interval), the effect size (Hedge’s g), and the statistical power (Stat Power).  

In addition to standard hypothesis testing results, we have further computed the effect size (Hedge’s g) and the statistical power to provide a more comprehensive evaluation of the robustness of our conclusions \citep{cohen2023statistical}. Specifically, Hedge’s g is a standardized measure of mean difference between two groups, which is an improved version of Cohen’s d designed for small sample sizes. It is important to note that effect size is independent of sample size, making it a more objective indicator of the generalizability of our findings, even for small samples. Statistical power (Stat Power), on the other hand, represents the probability of correctly rejecting the null hypothesis — i.e., detecting a true effect when one exists. A high statistical power reduces the risk of Type II errors (failing to detect a real effect).  

As shown in Table \ref{tab:statistic_testing}, at a 95\% confidence level, the use of ORLM significantly improves both solution time and accuracy across all populations (total, expert, and student samples). The confidence intervals suggest that, after implementing ORLM, accuracy improves by approximately 10\%–25\%. Additionally, the total time saved in solving optimization modeling problems is 1.8-2.2 hours, with experts saving approximately 1.5–2 hours and students saving around 2–2.5 hours. Additionally, the absolute value of Hedge’s g values exceed 1, demonstrating a substantial effect size and indicating a large difference between the experimental and control groups. Furthermore, the statistical power values all exceed 0.8, which is considered a strong threshold, ensuring a high probability of detecting a true effect and minimizing the risk of a Type II error.  

In conclusion, our findings provide compelling evidence that ORLM significantly enhances both accuracy and efficiency for both students and experts in solving operational research modeling problems.
}

\subsection{Examples of Questions with Different Difficulty Levels}
\label{app:compare_easy_and_hard}
\subsubsection{Easy Example}
\label{app: easy example}
A company sells custom scooters and bikes for customers. The profit per scooter is \$200 and the profit per bike is \$300. Each product requires time with the design team and engineering team. Each scooter needs 2 hours with the design team and 3 hours with the engineering team. Each bike needs 4 hours with the design team and 5 hours with the engineering team. Per month, there are 5000 hours available on the design team and 6000 hours available on the engineering team. How many of each should the company make per month to maximize profit?

\subsubsection{Medium Example}
\label{app: medium example}

We consider the following employee scheduling problem: A company needs to reassign 6 employees to complete the work over the next ten days. The daily workforce requirements and the availability of each employee for reassignment are known. Let \( w_j \) denote the number of employees required on day \( j \), and \( \alpha_{ij} \) indicate whether employee \( i \) is available for reassignment on day \( j \). Furthermore, let \( \mathcal{I} \) represent the set of all employees and \( \mathcal{J} \) denote the set of working days.

The objectives are twofold: (1) to ensure that each day’s workforce requirements are met as closely as possible, minimizing the number of unfulfilled positions; and (2) to balance the distribution of workdays among employees, minimizing disparities in their workloads.

\subsubsection{Hard Example}
\label{app: hard example}

The scheduling of hot coil transportation involves vehicle dispatch between warehouses and between warehouses and docks. Transportation tasks between warehouses are called \textit{transfer tasks}, while those between warehouses and docks are called \textit{dock tasks}.

Before the start of each shift, schedulers need to assign vehicles to the existing steel coil transportation tasks, determine the execution time for each task, and ensure all tasks are assigned. Tasks have merging rules: tasks with the same starting and ending points can be executed by the same vehicle, but the total weight and the total number of steel coils must not exceed the vehicle's limits. Vehicles do not pick up new tasks while en route; they can only pick up a new task after completing the current one.

\textbf{Task Format:} Steel coil ID, steel coil weight, starting warehouse, destination warehouse (dock), ship ID, task priority

\begin{enumerate}
    \item Minimize the number of vehicles used
    \item Ensure that all tasks are completed as early as possible
    \item Prioritize tasks with high priority
\end{enumerate}

Constraints is listed as follows:

\begin{enumerate}
    \item Vehicles have an initial parking spot and must start from this spot when executing the first task of the shift
    \item The number of steel coils loaded on a vehicle must not exceed the vehicle's limit
    \item The weight of steel coils loaded on a vehicle must not exceed the vehicle's limit
    \item The vehicle's transportation speed must be within the maximum and minimum speed limits
    \item Different ships at the same dock must be loaded sequentially; the next ship's loading can only start after the previous ship's loading is completed
    \item There is a limit to the number of vehicles simultaneously executing dock tasks
    \item There is a limit to the number of vehicles simultaneously executing transfer tasks
    \item The number of vehicles operating simultaneously in the warehouse area has an upper limit
    \item No new tasks should be assigned to vehicles in the last half hour of the current shift
\end{enumerate}

\noindent The sequence of tasks and estimated time nodes for all vehicles within this shift.

\end{APPENDICES}

\end{document}